\documentclass[a4paper,conference]{IEEEtran}
\IEEEoverridecommandlockouts
\usepackage{cite}
\usepackage{amsmath,amssymb,amsfonts}
\usepackage{algorithmic}
\usepackage{graphicx}
\usepackage{textcomp}
\usepackage{xcolor}
\def\BibTeX{{\rm B\kern-.05em{\sc i\kern-.025em b}\kern-.08em
    T\kern-.1667em\lower.7ex\hbox{E}\kern-.125emX}}

\usepackage{mathtools}
\DeclarePairedDelimiter\floor{\lfloor}{\rfloor}

\usepackage{array}
\usepackage{url}

\usepackage{color}

\usepackage{mathtools}

\newcommand{\eqsize}{\small}
\newcommand{\GG}{\mathcal{G}}
\newcommand{\SSS}{\mathcal{S}}
\newcommand{\PP}{\mathbf{P}}
\newcommand{\sa}{S_i^{a\rightarrow \{\mathbb{N}^{2}\}}}
    
\begin{document}

\title{Generalized Shortest Path-based Superpixels for Accurate  Segmentation of Spherical Images}

\author{\IEEEauthorblockN{R{\'e}mi Giraud, Rodrigo Borba Pinheiro, Yannick Berthoumieu}
\IEEEauthorblockA{\textit{Bordeaux INP, Univ. Bordeaux, CNRS, IMS, UMR 5218} \\
F-33400 Talence, France \\
remi.giraud@ims-bordeaux.fr, rodrigo.borba\_pinheiro@bordeaux-inp.fr, yannick.berthoumieu@ims-bordeaux.fr}
}

\maketitle

\begin{abstract}
Most of existing superpixel methods are designed to segment standard planar images as pre-processing for computer vision pipelines. 
Nevertheless, the increasing number of applications based on wide angle capture devices,
mainly generating 360$^\text{o}$ spherical images, 
have enforced the need for dedicated superpixel approaches.
In this paper, we introduce a new superpixel method for spherical images called SphSPS (for Spherical Shortest Path-based Superpixels).
Our approach respects the spherical geometry and
generalizes the notion of shortest path between a pixel and a superpixel center 
on the 3D spherical acquisition space.
We show that the feature information on such path can be 
efficiently integrated into our clustering framework 
and jointly improves the 
respect of object contours and the shape regularity.
To relevantly evaluate this last aspect in the spherical space, 
we also generalize a planar global regularity metric.
Finally, the proposed SphSPS method obtains significantly better performance than both planar and recent spherical superpixel approaches 
on the reference
360$^\text{o}$ spherical panorama segmentation dataset.
\end{abstract}


\section{Introduction}

The growing in resolution and quantity of image data     
%
%
has highlighted the need for efficient under-representations
to reduce the computational load of computer vision pipelines. 
%
In this context, 
superpixels were popularized with \cite{achanta2012}
to reduce the image domain to  
irregular regions having approximately the same size
and homogeneous colors. 
%
Contrary to regular multi-resolution schemes, 
a result at the superpixel scale can be very close to the optimal 
one at the pixel scale.
Superpixels have been successfully used in many 
applications such as: 
semantic segmentation \cite{tighe2010,wang2013med},
optical flow estimation \cite{menze2015object} 
or style transfer \cite{liu2017photo}.
The main issue to deal with 
is the 
irregularity between all regions that may
prevent from using the standard neighborhood-based tools. 
Nevertheless, this issue has been addressed 
in graph-based approaches \cite{gould2014}, 
using 
neighborhood structure \cite{giraud2017_spm}, 
or 
within
deep learning frameworks \cite{liu2018learning}.

At the same time, the use of new acquisition devices
capturing wide angles, 
such as fish eyes, 
generally covering a 360$^\text{o}$ field of view 
has become more and more popular.
These devices offer a global capture of the environment, 
particularly interesting 
for applications such as autonomous driving.
With 
a depth-aware system, 
the 
intensity can be projected on a 3D point cloud.
Otherwise, the image sphere is generally projected on a discrete 2D plane 
to generate an equirectangular image inducing distortions \cite{zorin1995correction}.
In this context, several works, \emph{e.g.}, \cite{cabral2014piecewise,sakurada2015change}
have used standard planar superpixels 
although they 
do not consider the geometry distortions in the equirectangular image, 
that may limit the segmentation accuracy and 
their interpretation on the spherical acquisition space \cite{zhao2018}.
%

Many superpixel approaches have been proposed over the years, 
most exclusively to segment standard planar images.
These methods use 
watershed \cite{machairas2015},
region growing \cite{levinshtein2009}, 
eikonal-based \cite{buyssens2014},
graph-based energy \cite{liu2011},
or even coarse-to-fine algorithms \cite{yao2015}.
A significant breakthrough was obtained with the SLIC method \cite{achanta2012},
locally adapting a $K$-means algorithm
on a trade-off between distances in the spatial and CIELab color space to generate superpixels.
The method has few parameters and a low processing time, 
but may struggle to jointly capture object borders 
and provide regular shapes.
Many improvements of SLIC have been proposed using 
boundary constraint \cite{zhang2016},
advanced feature space \cite{chen2017},
non-iterative clustering \cite{achanta2017superpixels},
a shortest path approach \cite{giraud2018_scalp},
or even deep learning processes 
\cite{liu2018learning}
%
although these last methods present 
the usual 
 limitations, in terms of resources, training time, large dataset needed, and
applicability to other images.

For spherical images, 
The unsupervised segmentation approach of \cite{felzenszwalb2004}
has been extended 
in \cite{yang2016efficient}, 
but generates very irregular regions, 
not considered as superpixels.
More recently, the SLIC method was extended to produce 
spherically regular superpixels \cite{zhao2018}.
Pixels are projected on the unit sphere for computing the spatial constraints
and produce regular superpixels in the spherical space.
Besides the display interest, the respect of the acquisition space geometry
enables to more accurately segment the image objects \cite{zhao2018}.
Nevertheless, this approach comes with the same limitations as SLIC,
\emph{i.e.}, 
limited adaptability to different contexts
with severe non robustness to textures or noise
due to the use of a standard color feature space,
and no explicit integration of contour information. 
These limitations are addressed in \cite{giraud2018_scalp}, 
for which authors obtain significantly higher accuracy 
for standard planar image segmentation
by considering the color and contour features along the shortest path between the pixel and the superpixel.
%
%

%



 \subsubsection*{Contributions}
%
In this paper, we address the limitations of the spherical approach 
of \cite{zhao2018}, 
by proposing in Section \ref{sec:sps} a new superpixel method called
SphSPS (Spherical Shortest Path-based Superpixels).
SphSPS is based on the same spherical $K$-means approach of \cite{zhao2018} but
exploits more advanced features \cite{chen2017} and 
generalizes the notion of shortest path \cite{giraud2018_scalp},
to the acquisition space, here the spherical one.
To this end, a dedicated fast shortest path algorithm is defined
to integrate the information of this large number of pixels 
into the method.

\begin{figure}[t]
\centering
{\scriptsize
\begin{tabular}{@{\hspace{1mm}}c@{\hspace{4mm}}c@{\hspace{1mm}}}
\includegraphics[width=0.225\textwidth,height=0.112\textwidth]{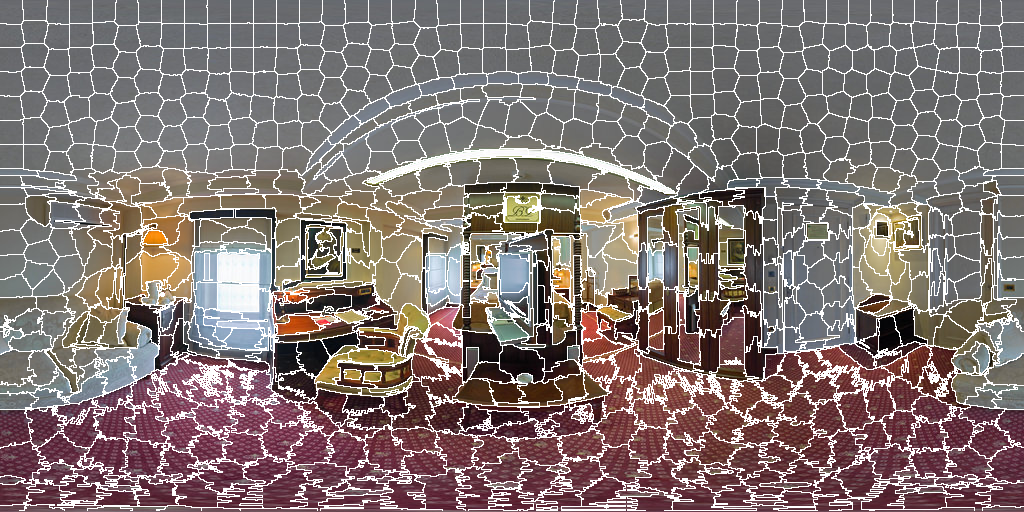}&
\includegraphics[width=0.1125\textwidth,height=0.112\textwidth]{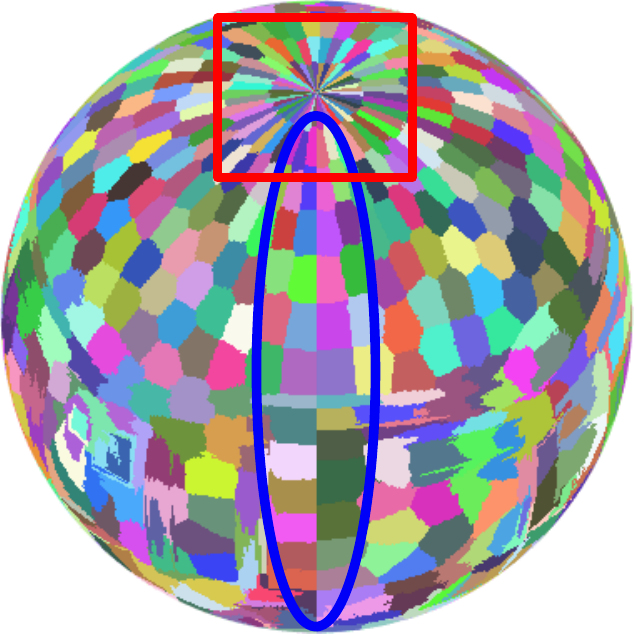}\\
\multicolumn{2}{c}{Standard planar superpixels using \cite{chen2017}}\\[0.225ex]
\includegraphics[width=0.225\textwidth,height=0.112\textwidth]{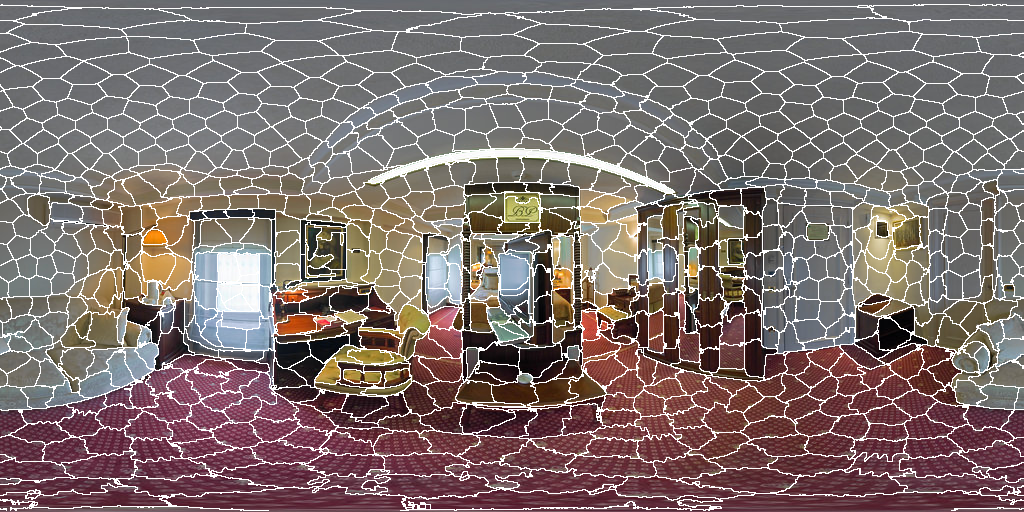}&
\includegraphics[width=0.1125\textwidth,height=0.112\textwidth]{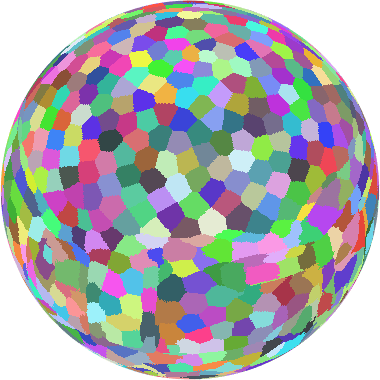}\\
\multicolumn{2}{c}{Spherical superpixels using the proposed SphSPS method}\\
\end{tabular}}%
\caption{Example of superpixel segmentation with a planar \cite{chen2017} and the proposed spherical SphSPS method on a 360$^\text{o}$ panorama image. 
SphSPS provides accurate superpixels that are
regular in the spherical acquisition space (red square) and
connected at horizontal boundaries (blue ellipse).}%
\label{fig:sps_ex_intro}
\end{figure}

%
%
%
%
SphSPS generates in very limited processing time accurate and regular spherical superpixels
(see Figure \ref{fig:sps_ex_intro}).
To relevantly evaluate the regularity aspect in the spherical space, 
we also propose a generalization of the global regularity measure \cite{giraud2017_jei} (Section \ref{sec:grs}).
SphSPS
 obtains higher segmentation performance than the state-of-the-art methods
 on the 
reference 360$^\text{o}$ spherical panorama segmentation dataset \cite{wan2018} (Section \ref{sec:results}).%


\section{Spherical Shortest Path-based Superpixels}
\label{sec:sps}

To introduce SphSPS,
we first present 
the {$K$-means} method \cite{achanta2012} (Section \ref{subsec:slic}) and
its spherical adaptation \cite{zhao2018} (Section \ref{subsec:spherical_geometry}).
Then, we present the feature extraction method on a  planar shortest path \cite{giraud2018_scalp} (Section \ref{subsubsec:planar_path})
and our generalization to the spherical space (Sections \ref{subsubsec:gen_path} and \ref{subsec:sps_path}).

\subsection{Planar K-means Iterative Clustering\label{subsec:slic}}

SphSPS is based on SLIC (Simple Linear Iterative Clustering) \cite{achanta2012} 
using an iteratively constrained $K$-means clustering 
of pixels.
Superpixels $S_i$ are first initialized
as blocks of size $s{\times}s$,
described by the average CIELab colors $C_{S_i}$ 
and barycenter position $X_{S_i}=[x_i,y_i]$ of pixels in $S_i$.
The clustering 
for each pixel $p$
relies on a
color  $d_c(p,{S_i})={\|C_p-C_{S_i}\|}_2^2$, 
and a spatial distance $d_s(p,{S_i})={\|X_p-X_{S_i}\|}_2^2$.
At each iteration,
each superpixel $S_i$ is compared to all pixels $p=[C_p,X_p]$, 
of color $C_p$ at position $X_p$, 
within a $(2s$$+$$1)$${\times}$$(2s$$+$$1)$ area $A_i$ around its barycenter $X_{S_i}$.
A pixel $p$ is associated to the superpixel minimizing the distance $D$ defined as: \vspace{-0.225cm}

{\eqsize
\begin{equation}
D(p,S_i)=d_c(p,{S_i}) + d_s(p,{S_i})\frac{m^2}{s^2} ,  \label{slic} 
\end{equation}
}%
with $m$, the trade-off parameter setting the shape regularity. 
Finally, a post-processing step 
ensures the region connectivity.



\subsection{Spherical Geometry\label{spherical_geometry}\label{subsec:spherical_geometry}}

In the spherical acquisition space,
vertical and horizontal
coordinates are respectively projected 
to the meridians and circles of latitude,
so the spherical image has a width twice superior to its height.
SphSPS is based on the same adaptation of the planar $K$-means 
method to the spherical geometry as \cite{zhao2018}, 
that requires three steps.
The first one is the initialization of the $K$ superpixels. 
To spread the barycenters along the sphere, we also use the Hammersley sampling \cite{wong1997sampling}.
The second step is the search area that must consider the proximity of pixels in the spherical space.
For instance, superpixels on the image top and bottom have larger search areas.
This area $A_i$, is defined for each superpixel $S_i$ of barycenter $X_{S_i}=[x_i,y_i]$ as: \vspace{-0.2cm}

{\eqsize
\begin{align}
\hspace{-0.05cm} A_i = \hspace{-0.05cm}\{[x,y] | x_i\hspace{-0.05cm}-\hspace{-0.05cm}\frac{S}{\text{sin} \phi}\leq \hspace{-0.05cm}x\hspace{-0.05cm}\leq x_i+\frac{S}{\text{sin} \phi} , y_i\hspace{-0.05cm}-\hspace{-0.05cm}S \leq y \leq y_i\hspace{-0.05cm}+\hspace{-0.05cm}S\},
\end{align}
}%
\noindent with $\phi=y\pi/h$ the polar angle corresponding to the $y$-th row for an image
of height $h$ and width $w$, and the average superpixel size $S=w/\sqrt{K\pi}$.
The 360$^\text{o}$ geometry aspect must also be handled
to horizontally connect the pixels. 
This is done with a left/right warping when the search region falls outside the image boundaries \cite{zhao2018}.
The third aspect is the computation of the spatial distance, 
that must also be done 
in the spherical space.
For each image pixel $X=[x,y]$ 
the projection on the 3D 
acquisition space $X^a=[x^a,y^a,z^a]$ is computed as: \vspace{-0.15cm}

%
%
%
{\eqsize
\begin{equation}
\hspace{-0.25cm}
\begin{array}{l}
\left\{\hspace{-0.2cm}
\begin{array}{ll}
x^a = \text{sin}(\frac{y\pi}{h})\text{cos}(\frac{2x\pi}{w}) \\[0.25ex]
y^a = \text{sin}(\frac{y\pi}{h})\text{sin}(\frac{2x\pi}{w}) \\[0.25ex]
z^a = \text{cos}(\frac{y\pi}{h})
 \end{array} 
 \right.
  \hspace{-0.1cm} 
\leftrightarrow \hspace{0.1cm} 
\left\{\hspace{-0.2cm}
\begin{array}{ll}
x=\hspace{-0.00cm} \floor{\frac{\text{arctan2}({y^a,x^a})w}{2\pi}} \\[-0.5ex]
\\[-0.25ex]
y=\hspace{-0.00cm} \floor{\frac{\text{arccos}(z^a)h}{\pi}} \\
 \end{array} \hspace{-0.35cm}
 \right.%
 \end{array}.   \label{xy_xyz}
\end{equation}
}%
%
%
Note that $x\in [-\frac{w}{2},\frac{w}{2}]$, when computed from $X^a$, 
so we map $x$ on the image domain  
with $x \leftarrow x+w$, if $x\leq0$.

The most straightforward 3D spatial distance 
is the Euclidean one 
$d_s(X_p^a,X_{S_i}^a)=\|X^a_p - X^a_{S_i}\|_2^2$.
SphSPS uses the spherical and computationally costless cosine dissimilarity distance proposed in \cite{zhao2018} 
as 
$d_s(X_p^a,X_{S_i}^a)=1-\left< X_p^a, X_{S_i}^a\right>$.
%
Note that 
with adjusted parameter $m$ \eqref{slic}, 
both distances can achieve almost similar 
performances for \cite{zhao2018} (see Section \ref{sec:results}).

\subsection{Generalized Shortest Path Method}

\subsubsection{Feature extraction on a shortest path\label{subsubsec:planar_path}}

In \cite{giraud2018_scalp}, color and contour information 
of pixels $q$
on the planar shortest path $\mathbf{P}_{p,S_i}$ between a pixel $p$ 
and a superpixel $S_i$
are used to improve segmentation accuracy and regularity.
%
SphSPS also integrates these features and has the same clustering distance $D$ than \cite{giraud2018_scalp}.
Nevertheless, in the following, the shortest path $\mathbf{P}_{p,S_i}$  differs since we compute it in the spherical space. 
%
%
%

%
First, 
to relevantly increase the regularity 
and prevent non-convex shapes from appearing, 
the color distance of the pixels on the path 
is added to the color distance $d_c$ such that: \vspace{-0.2cm}

{\eqsize
\begin{equation}
d_c(p,S_i,\PP_{p,S_i})\hspace{-0.05cm}=\hspace{-0.05cm}\lambda d_c(p,S_i)\hspace{-0.1mm}+\hspace{-0.1mm}
\frac{1-\lambda}{|\PP_{p,S_i}|}\hspace{-0.1cm}\sum_{q\in \PP_{p,S_i}}\hspace{-0.1cm}d_c(q,S_i),   
\label{path_color}
\end{equation}
}%
\noindent
with $\lambda$ a trade-off parameter usually set to $0.5$. 
                 
The contour information can also be considered to increase
the respect of objects borders using any contour map $\mathcal{C}$, with values between 0 and 1.
A contour term $d_\mathcal{C}$ is defined as:  \vspace{-0.15cm}

{\eqsize
\begin{equation}
d_{\mathcal{C}}(\PP_{p,S_i}) = 1 + \gamma \hspace{0.1cm} \underset{q\in \PP_{p,S_i}}{\text{max}}\hspace{0.05cm}\mathcal{C}(q) , \label{path_contour} 
\end{equation}
}%
\noindent
with $\gamma$ the parameter penalizing the crossing of a contour.

The final clustering distance of SphSPS
is defined as:   \vspace{-0.2cm}

{\eqsize
\begin{equation}
   D(p,S_i)=\left(d_c(p,S_i,\PP_{p,S_i}) + d_s(p,S_i)\frac{m^2}{s^2}\right)d_{\mathcal{C}}(\PP_{p,S_i})  , \label{newdist} 
\end{equation}}%
\noindent with the spherical spatial distance $d_s$ using the cosine dissimilarity as
 $d_s(X_p^a,X_{S_i}^a)=1-\left< X_p^a, X_{S_i}^a\right>$ \cite{zhao2018},
 and $\PP_{p,S_i}$ the proposed spherical shortest path computed as follows.

\subsubsection{Generalized shortest path\label{subsubsec:gen_path}} 
\begin{figure}[t]
\centering
\includegraphics[width=0.435\textwidth,height=0.1775\textwidth]{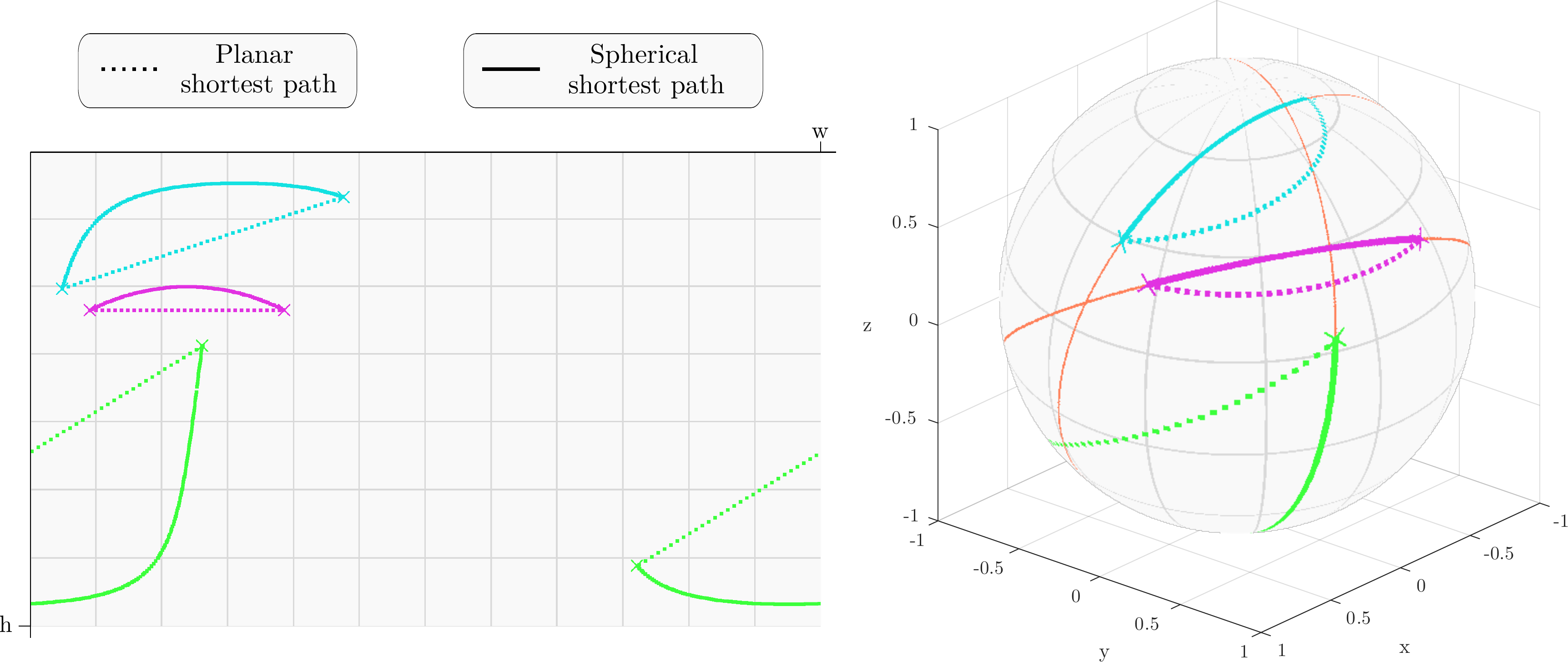}
\caption{Examples of planar (dotted lines) and spherical shortest path (full lines) between points in the 2D image space (left) and 3D acquisition space (right).
The spherical path follows the shortest geodesic path on the sphere.
}%
\label{fig:spherical_path}
\end{figure}

In Figure \ref{fig:spherical_path}, 
we compare shortest paths in the planar space,
as in \cite{giraud2018_scalp},
and in the spherical one as in SphSPS.
%
%
%
With planar images, 
since no distortions are introduced between the
acquisition and the image space ($\mathbb{N}^2$),
they are considered equivalent.
Hence, 
the shortest path reduces to a linear path 
and can be easily computed with a discrete algorithm \cite{bresenham1965}.
%
%
%
Nevertheless, in general, the shortest path should be computed in the
acquisition space, than can be spherical or even circular using fisheyes with different capture angles.
Hence, the generalized formulation of the shortest path problem 
computes it 
in the acquisition space ($\mathbf{P}_{p,S_i}^a$) and projects it back to the planar image space:
\vspace{-0.15cm}

{\eqsize
\begin{equation}
 \mathbf{P}_{p,S_i}  = \mathbf{P}_{p,S_i}^a \hspace{0.25cm} \overrightarrow{\text{\small proj}} \hspace{0.25cm} \{\mathbb{N}^2\} . \label{proj_path}
\end{equation}
}%


\subsubsection{Shortest path in the spherical space\label{subsec:sps_path}}
%
%
The spherical shortest path consists in following the geodesic along the sphere \cite{gromov1983filling},
lying on a \textit{great circle} 
(in orange color in Figures \ref{fig:spherical_path} and \ref{fig:great_circle}), 
containing the two points and the sphere center. 
Tangential methods to extract way-points on the great circle 
have been formalized for instance in \cite{karney2013algorithms}.
Nevertheless, such theoretical approaches
use many trigonometric computations 
that impact the performance.
In the following we propose a simpler reformulation of the spherical geodesic path problem.
%

 \smallskip

\subsubsection*{Fast geodesic path implementation}

\begin{figure}[t!]
\centering
{\footnotesize
\includegraphics[width=0.385\textwidth,height=0.1775\textwidth]{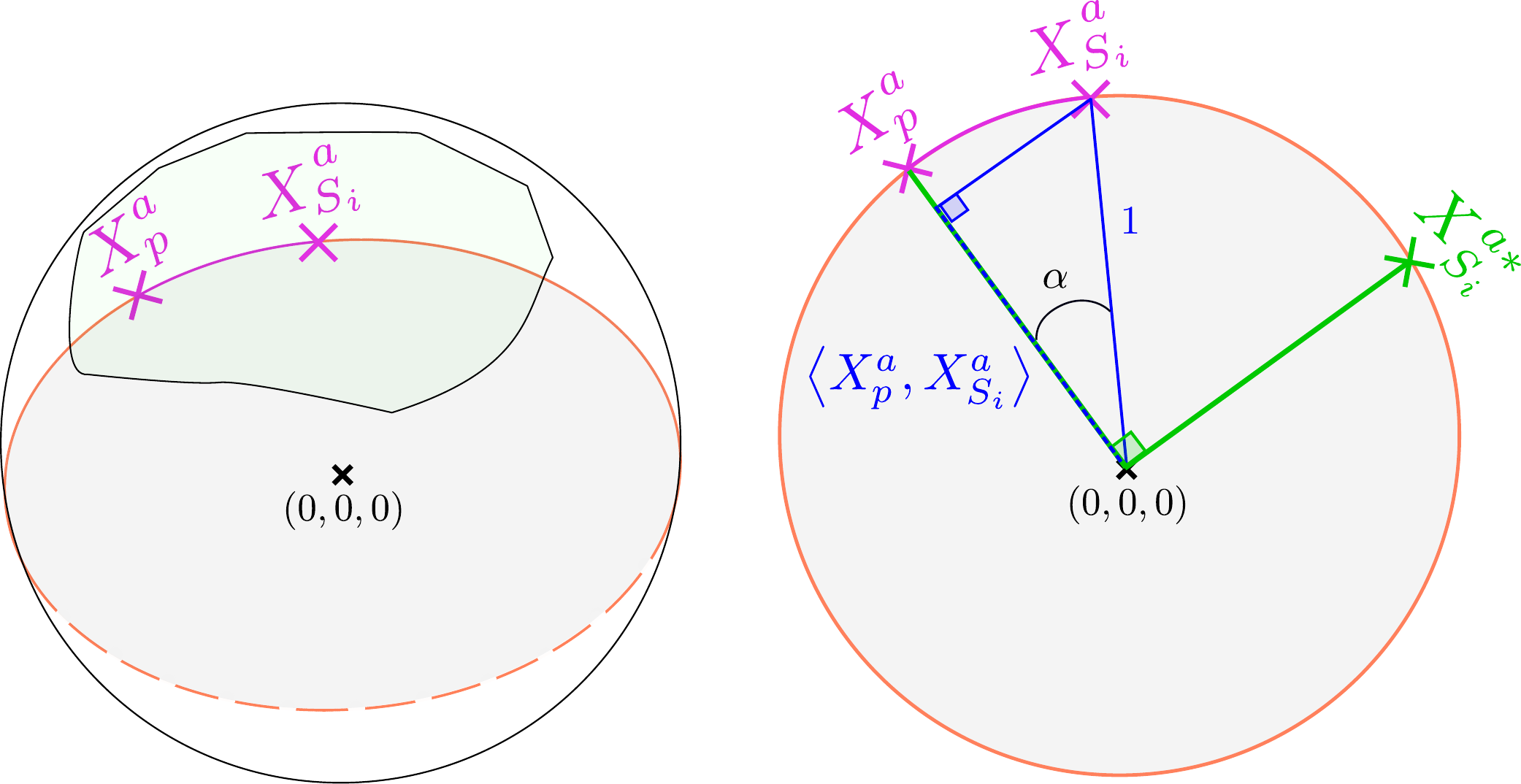}}%
\caption{Computation of the spherical shortest path.
The orthogonal coordinate system $[\vec{X_p^a},\vec{X_{S_i}^{a*}}]$ is computed from projection of  $X_{S_i}^a$ on $X_p^a$ \eqref{co_system}.
The  angle  $\alpha$ between the positions is then
is used to sample 3D points on the path \eqref{sph_path}.%
%
}%
\label{fig:great_circle}
\end{figure}

For each comparison of a pixel at $X_p^a$ to a superpixel of barycenter $X_{S_i}^a$,
we propose to first compute 
an orthogonal coordinate system $[\vec{X_p^a},\vec{X_{S_i}^{a*}}]$
within their great circle.
To build such system, we 
perform an orthogonalization process
to get the position ${X_{S_i}^{a*}}$, 
creating an orthogonal vector to $X_p^a$ 
within the great circle such as: \vspace{-0.115cm}

{\eqsize
\begin{equation}
{X_{S_i}^a}^* = \frac{X_{S_i}^a - \left<X_p^a,X_{S_i}^a\right>X_p^a}{\left\|X_{S_i}^a - \left<X_p^a,X_{S_i}^a\right>X_p^a\right\|_2} , 
\label{co_system}
\end{equation}
}%
\noindent
with the scalar product $\left<X_p^a,X_{S_i}^a\right>$ already computed for the spatial distance $d_s$.
Then, the angle between the two points is simply obtained with 
$\alpha = \text{arccos}\left(\left<X_p^a,X_{S_i}^a\right>\right)$.
Finally, 
the geodesic path $\mathbf{P}_{p,S_i}^a$ is defined
within 
$[\vec{X_p^a},\vec{X_{S_i}^{a*}}]$, by starting from the pixel position, 
and linearly increasing the angle shift from $0$ to $\alpha$, 
to reach the superpixel barycenter such as: \vspace{-0.175cm}
%
%
%

{\eqsize
    \begin{equation}
     \mathbf{P}_{p,S_i}^a = \text{cos}(\mathbf{\alpha_N})X^a_p + \text{sin}(\mathbf{\alpha_N})X^{a*}_{S_i} ,  \label{sph_path}
    \end{equation}
}%
with $\mathbf{\alpha_N}=\frac{[0, N-1]}{N-1} \alpha\in\mathbb{R}^N$, intermediate angles 
to linearly sample $N$ points between the two positions.
The geodesic path 
is finally projected in the planar space \eqref{xy_xyz} to get $\mathbf{P}_{p,S_i}$ \eqref{proj_path}.
By this way, we obtain the shortest spherical path coordinates with simple calculations, dividing the processing time by a factor 2 compared to tangential approaches.
An example of spherical shortest path on a great circle with the the computation of the corresponding coordinate system is illustrated in Figure \ref{fig:great_circle}.

\vspace{0.05cm}

\subsubsection*{Optimization} 

First, for each superpixel, we can
store the color distance 
computed to each tested pixel,
reducing the processing time by $50\%$.
Then, contrary to the planar linear path algorithm \cite{bresenham1965}
we can exploit path redundancy.
If the path of a pixel to a superpixel crosses
a previously computed path to the same superpixel, 
the rest of the path should be the same
since they lie on the same great circle.
So we can also store the average color and contour information 
on the path for each crossed pixel.
This is done efficiently using recursive implementation. 
By this way, for many pixels we are able to directly access 
the large quantity of information contained in the shortest path, 
again reducing the processing time by $50\%$.

\section{Generalized Global Regularity Measure}
\label{sec:grs}

Superpixels tend to optimize a color and spatial trade-off,
so metrics should mainly evaluate 
object segmentation and regularity performances.
This last aspect has rarely been evaluated 
although most methods 
 have a regularity parameter that may significantly impact 
 superpixel-based pipelines.
 Moreover, 
the standard  compactness metric \cite{schick2012}, 
which is the only one extended to the spherical space \cite{zhao2018}
was proven very limited \cite{giraud2017_jei}.
In this section, we propose a 
new way to relevantly evaluate the regularity in the acquisition space.

\subsection{Limitation of the Compactness Measure}

In \cite{zhao2018}, the compactness measure COM \cite{schick2012} 
is extended to the spherical case.
The regularity of a segmentation $\SSS=\{S_i\}$
is only seen as a notion of circularity, 
computed as: 
 \vspace{-0.05cm}

{\eqsize
\begin{equation}
\text{COM} = \frac{1}{\sum\limits_{S_i\in \mathcal{S}} |S_i|}\sum\limits_{S_i\in \mathcal{S}} Q(S_i)|S_i| ,  \vspace{-0.0cm}
\label{circu}
\end{equation}
}%
with $Q(S_i)$=$(4\pi |S_i| - |S_i|^2)/|P(S_i)|^2$ the spherical
isoperimetric quotient using $P(S_i)$ the perimeter of $S_i$ in the spherical space \cite{osserman1978isoperimetric}. 
Hence, 
each superpixel is independently compared to a circular shape, 
such that for instance, ellipses can have higher COM measures than squares.
In \cite{giraud2017_jei}, this metric has been proven 
highly sensitive to boundary noise and inconsistent with the superpixel size.
%
Moreover, in \cite{zhao2018} it even fails to differentiate spherical and planar-based methods.

\begin{figure}[t]
\centering
\includegraphics[width=0.46\textwidth,height=0.345\textwidth]{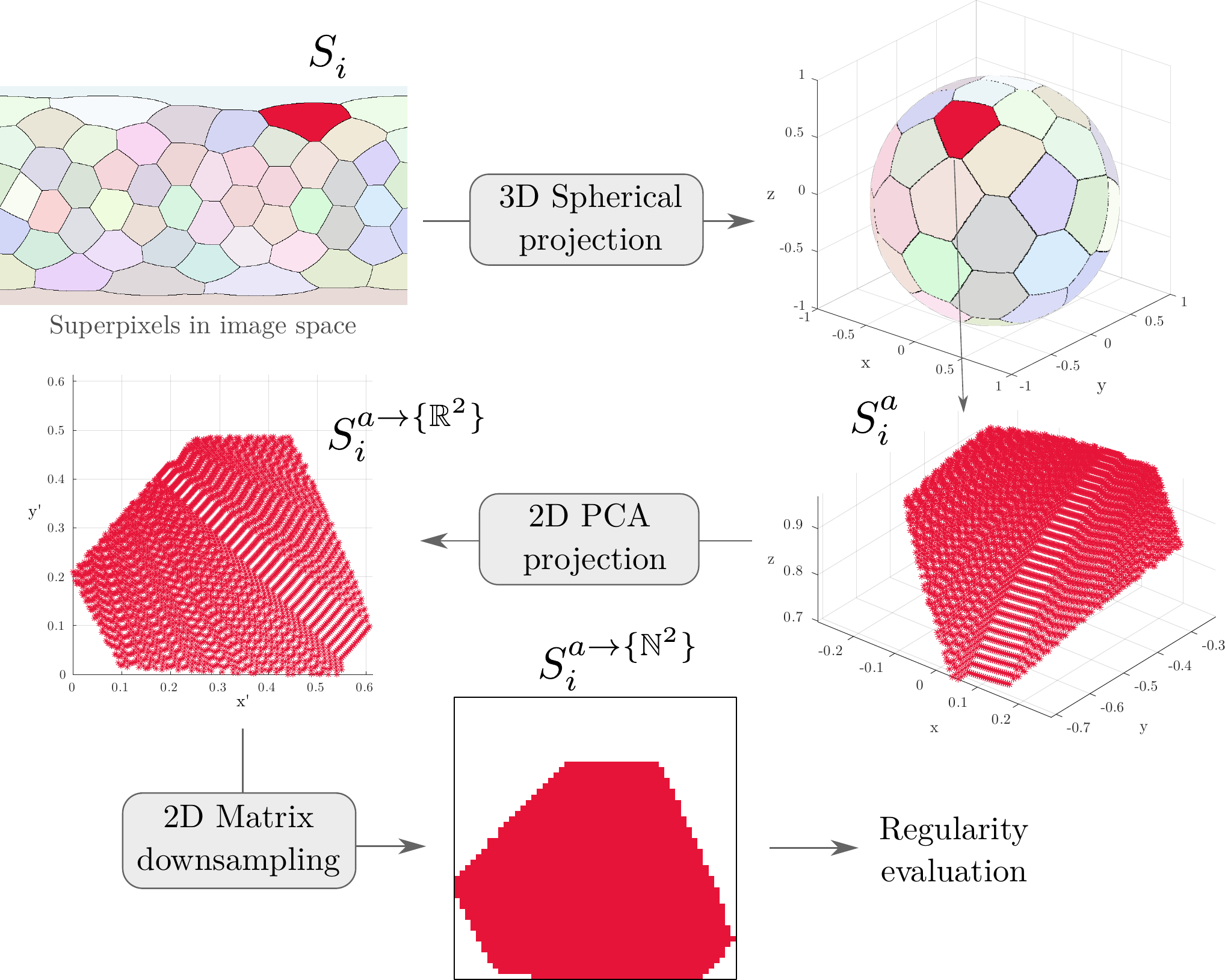}
\caption{Illustration of the projection process of the proposed 
Generalized Global Regularity (G-GR) metric \eqref{grs}.
A superpixel shape $S_i$ is projected in the acquisition space ($S_i^a$), then 
on a two dimensional one using a PCA  ($S_i^{a\rightarrow \{\mathbb{R}^2\}}$),
then downsampled to generate a 2D matrix ($S_i^{a\rightarrow \{\mathbb{N}^2\}}$), 
allowing for instance to compute a convex hull to measure its regularity.}%
\label{fig:grs}
\end{figure}

\subsection{Generalized Global Regularity Metric}

\subsubsection{Global regularity metric}

In \cite{giraud2017_jei}, a global regularity metric (GR) is introduced, to address the issues of the compactness.
%
First, the Shape Regularity Criteria (SRC) is defined 
to robustly evaluate the convexity, the contour smoothness, and the 2D balanced repartition of each superpixel. 
Convexity and smoothness properties are computed with respect to the discrete convex hull containing the shape.
%
%

As for the compactness COM \eqref{circu},
SRC is independently computed for each superpixel, 
so \cite{giraud2017_jei} also introduces a Smooth Matching Factor (SMF) 
to evaluate the consistency of superpixel shapes.
Each superpixel is compared,
after registration on its barycenter,
to the average superpixel shape, created from
the superposition of all registered superpixels.


Finally, the notion of regularity is defined by the GR (Global Regularity) metric 
combining these two metrics such that: \vspace{-0.1cm}

{\eqsize
\begin{equation}
\text{GR}(\SSS) = \frac{1}{\sum\limits_{S_i\in \mathcal{S}} |S_i|}
\sum\limits_{S_i\in \mathcal{S}} |S_i|\text{SRC}(S_i)\text{SMF}(S_i) . \label{gr} 
\end{equation}
}%

\subsubsection{Generalization in the acquisition space}
%
%
Ideally the regularity 
should be evaluated in the acquisition space.
In our context, $S_i$ in the spherical acquisition space
gives $S_i^a$, a set of 3D positions on the unit sphere \eqref{xy_xyz}.
GR being based on 
the computation of convex hull, 
and
barycenter registration, 
it cannot be directly applied to such point clouds in $\mathbb{R}^3$.

To generalize the metric,
we propose to simply project the 3D points of $S_i^a$ 
on a discrete 2D plan, and then apply the initial GR.
The whole process is illustrated in Figure \ref{fig:grs}.
To do so, we first project a superpixel $S_i$ in the discrete image space 
to its acquisition one, here to get a spherical point cloud $S_i^a$.
Then, we apply a Principal Component Analysis (PCA) on $S_i^a$, 
and project the points on its two most significant eigenvectors
to reduce to a 2D point cloud $S_i^{a\rightarrow \{\mathbb{R}^2\}}$.
Finally, 
a downsampling is performed to obtain a discrete 2D shape  $S_i^{a\rightarrow \{\mathbb{N}^2\}}$.
By this way, each superpixel shape has a relevant discrete projection 
in the acquisition space.
The proposed Generalized Global Regularity (G-GR) metric is defined as:\vspace{-0.285cm}

{\eqsize
\begin{equation}
 \hspace{-0.05cm} \text{G-GR}(\SSS) \hspace{-0.05cm} = \hspace{-0.05cm} \frac{\sum\limits_{S_i\in \mathcal{S}} \left|\sa\right|\text{SRC}(\sa)\text{SMF}(\sa)}
 {\sum\limits_{S_i\in \mathcal{S}} \left|\sa\right|} . \hspace{-0.5cm}
 \label{grs}  
\end{equation}
}%
%
With the proposed G-GR metric, 
a gap is now visible such that
no planar methods have higher regularity than spherical ones
for a given number of superpixels (Section \ref{subsec:soa}). 
%
%

\section{Results}
\label{sec:results}

\subsection{Validation Framework}

\subsubsection{Dataset}

We consider the Panorama Segmentation Dataset (PSD) \cite{wan2018},
containing 75 360$^\text{o}$ equirectangular images of $512{\times}1024$ pixels,
having between 115 and 1085 segmented objects with an average size of 1334 pixels.
These images are taken from the standard spherical dataset SUN360 \cite{xiao2012recognizing},
and accurate ground-truth segmentations are provided by \cite{wan2018}.



\smallskip

\subsubsection{\label{subsec:metrics}Metrics}

%
To relevantly evaluate SphSPS performances and compare to state-of-the-art methods, 
we use the superpixel metrics
recommended in \cite{giraud2017_jei},
for several superpixel numbers.
The main aspects to evaluate are
the object segmentation
and spatial regularity performances,
which is robustly evaluated in the acquisition space
with the proposed G-GR metric \eqref{grs}.

For the segmentation aspect, 
the standard measure is the Achievable Segmentation Accuracy (ASA) \cite{liu2011}, 
highly correlated to the Undersegmentation Error 
\cite{neubert2012} as shown in \cite{giraud2017_jei}.
The ASA measures the overlap of a superpixel segmentation $\SSS$ 
with 
the ground truth objects, denoted $\GG$, such as: \vspace{-0.10cm}

{\eqsize
\begin{equation}
 \text{ASA}(\SSS,\GG) = \frac{1}{\sum\limits_{S_i \in \SSS}|S_i|}\sum_{S_i}\underset{G_j\in \GG}{\max}|S_i\cap G_j|.  \label{asa}
\end{equation}
}%
%
%
The Boundary-Recall (BR) is a commonly employed metric
to evaluate the detection of the ground truth contours $\mathcal{B(\GG)}$ 
by the boundaries of the superpixels $\mathcal{B}(\SSS)$ such that: \vspace{-0.15cm}

{\eqsize
\begin{equation}
\text{BR}(\SSS,\GG) = \frac{1}{|\mathcal{B}(\GG)|}\sum_{p\in\mathcal{B}(\GG)}\delta[\min_{q\in\mathcal{B}(\SSS)}\|p-q\|< \epsilon]  ,   \label{br}
\end{equation}
}%
\noindent with 
 $\epsilon$ a distance threshold set to $2$ pixels \cite{giraud2017_jei}, 
 and $\delta[a]=1$ when $a$ is true and $0$ otherwise.
%
%
%
%
To prevent methods generating  superpixels with fuzzy borders to get high performances \cite{giraud2017_jei}, 
BR results are compared to the Contour Density (CD), 
\emph{i.e.}, the number of pixels of superpixel borders.
%

The standard Precision-Recall curves can also be represented 
to illustrate the overall object contour detection  performances. 
These are computed on a contour probability map $\in [0,1]$
generated by averaging the superpixel borders obtained 
at different scales $K\in [50,3000]$. 
This map is thresholded by several intensities to get a binary contour map.
For each threshold, the Precision (PR), the percentage of accurate 
detection among the superpixel borders, 
is computed with the BR measure.
For all PR curves, to synthesize the contour detection performance, we also report
the maximum on all thresholds of the F-measure defined as: \vspace{-0.075cm}

{\eqsize
 \begin{equation}
 \text{F}=\frac{2\hspace{0.05cm}\text{PR}\hspace{0.05cm}\text{BR}}{\text{PR}+\text{BR}} . \label{fmeasure}
 \end{equation}}%

\subsubsection{Parameter settings}

SphSPS was implemented with MATLAB using C-MEX code, on a standard Linux computer with 
12 cores at 2.6 GHz with 64GB of RAM.
Contrary to \cite{zhao2018}, using the 3 average color features of the CIELab space, 
we use the 6 CIELab dimension space of \cite{chen2017}, also
including the features of neighboring pixels \cite{giraud2018_scalp}. 
%
%
In the shortest path, $N=15$ pixels are considered \eqref{sph_path}.
The number of iterations is set to 5, 
and the 
parameter $\lambda$ \eqref{path_color},
setting the trade-off between the central pixel and the ones on the shortest path,
is set to $0.5$ as in \cite{giraud2018_scalp}.
When used, the contour prior is computed from 
\cite{xie2015holistically} and $\gamma$ set to $10$ \eqref{path_contour}.
Finally, the parameter $m$ \eqref{newdist} is empirically set to $0.12$ to 
provide a visually satisfying trade-off between the respect of object contours and spatial regularity.
%
%

\newcommand{\wwh}{0.24\textwidth} 
\newcommand{\hhh}{0.2\textwidth}

\begin{figure}[t!]
\centering
{\scriptsize
\begin{tabular}{@{\hspace{0mm}}c@{\hspace{3mm}}c@{\hspace{0mm}}}
\includegraphics[width=0.22\textwidth,height=\hhh]{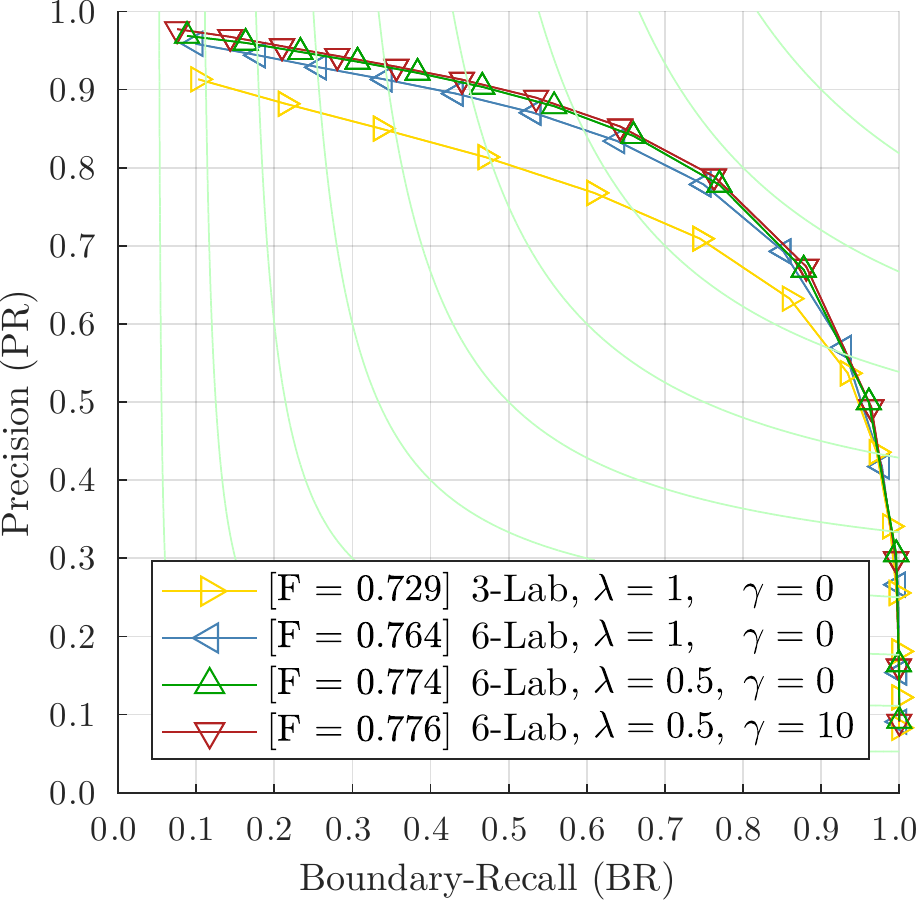}&
\includegraphics[width=\wwh,height=\hhh]{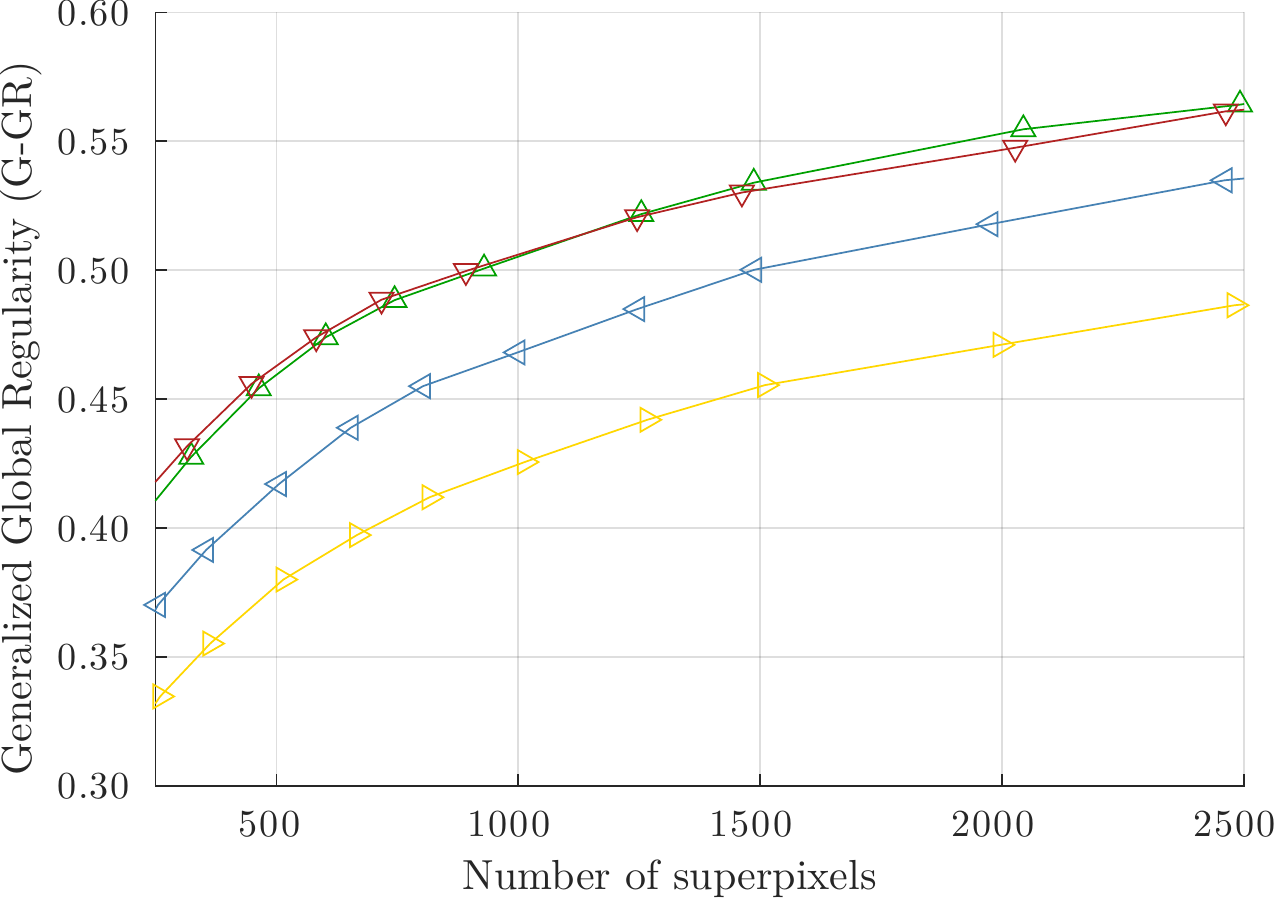}
\end{tabular}}
\caption{Impact of the SphSPS distance parameters. 
The contributions enable to significantly improve the
accuracy and regularity performances.}%
\label{fig:sps_param_curves}
\end{figure}

\newcommand{\gwwh}{0.15\textwidth} 
\newcommand{\whhh}{0.1\textwidth}

\begin{figure}[t!]
\centering
{\scriptsize
\begin{tabular}{@{\hspace{0mm}}c@{\hspace{1mm}}c@{\hspace{0mm}}}
\includegraphics[width=\gwwh,height=\whhh]{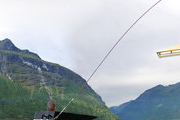}&
\includegraphics[width=\gwwh,height=\whhh]{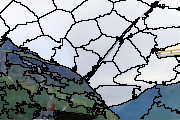}\\
 (a) Initial image & (b) 3-Lab, $\lambda$=$1,\gamma$=$0$\\[2ex]
\end{tabular}
\begin{tabular}{@{\hspace{0mm}}c@{\hspace{1mm}}c@{\hspace{1mm}}c@{\hspace{0mm}}}
\includegraphics[width=\gwwh,height=\whhh]{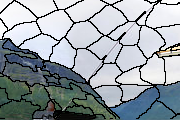}&
\includegraphics[width=\gwwh,height=\whhh]{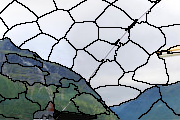}&
\includegraphics[width=\gwwh,height=\whhh]{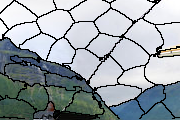}\\
(c) \textbf{6-Lab}$,\lambda$=$1,\gamma$=$0$ & 
(d) {6-Lab}$, $ {\color{black}\textbf{$\mathbf{\lambda}$=$\mathbf{0.5}$}}$,\gamma$=$0$ &
(e) {6-Lab}$,\lambda$=$0.5,$ {\color{black}\textbf{$\mathbf{\gamma}$=$\mathbf{10}$}}\\
\end{tabular}}%
\caption{Visual impact of  SphSPS parameters. Each contribution
relevantly increases the regularity and $\gamma=10$ 
integrates the contour prior information.
}%
\label{fig:sps_param_ex}
\end{figure}

\subsection{Impact of Contributions}

In this section, we show the impact of contributions within SphSPS.
We report for different distance settings 
the contour detection PR/BR curves, 
with the maximum F-measure \eqref{fmeasure}, 
and the regularity G-GR \eqref{grs} curves in Figure \ref{fig:sps_param_curves}, 
and a zoom on a segmentation example in Figure \ref{fig:sps_param_ex}.
With a 3 feature dimension space, 
SphSPS reduces to the spherical SLIC algorithm \cite{zhao2018}.
%
With the 6 dimension space, SphSPS uses the CIELab features of \cite{chen2017}, 
and the neighboring pixels information as in \cite{giraud2018_scalp},
with $\lambda=0.5$ the color distance \eqref{path_color}, 
and $\gamma=10$, the contour information  on the shortest path \eqref{path_contour}.

We demonstrate that each contribution improves the segmentation performance.
We can especially observe that 
the color distance on the shortest path, 
that strengthens the superpixel convexity and homogeneity,
indeed provides much more regular superpixels 
while also increasing the accuracy.

\subsection{\label{subsec:soa}Comparison with the State-of-the-Art Methods}


We compare the performances 
of the proposed SphSPS approach
to the ones of the 
state-of-the-art methods. 
We consider the 
planar methods
SLIC \cite{achanta2012},
LSC \cite{chen2017},
SNIC \cite{achanta2017superpixels} and
SCALP \cite{giraud2018_scalp}, 
and the spherical approach SphSLIC \cite{zhao2018}
in 2 different settings, \emph{i.e.}, considering the Euclidean (SphSLIC-Euc) and the 
Cosine (SphSLIC-Cos) distances (see Section \ref{spherical_geometry}). 
To ensure fair comparison, planar and SphSLIC-Euc methods are used with their default settings, 
since they provide a good trade-off between accuracy and regularity.
Note that for the SphSLIC-Cos method \cite{zhao2018},
results are reported 
for the regularity setting optimizing the segmentation accuracy,
since low  performance was obtained with default settings. 

\begin{figure*}[ht!]
\centering
{\scriptsize
\begin{tabular}{@{\hspace{0mm}}c@{\hspace{1mm}}c@{\hspace{1mm}}c@{\hspace{1mm}}c@{\hspace{0mm}}}
\includegraphics[width=0.22\textwidth,height=\hhh]{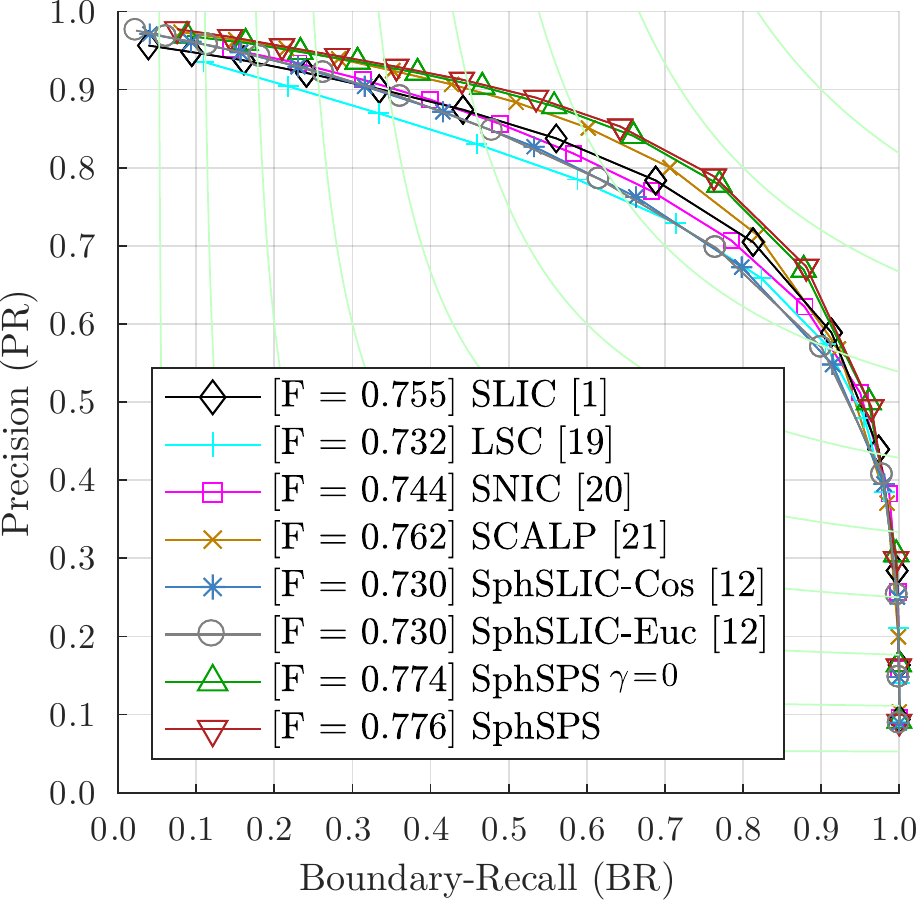}&
\includegraphics[width=\wwh,height=\hhh]{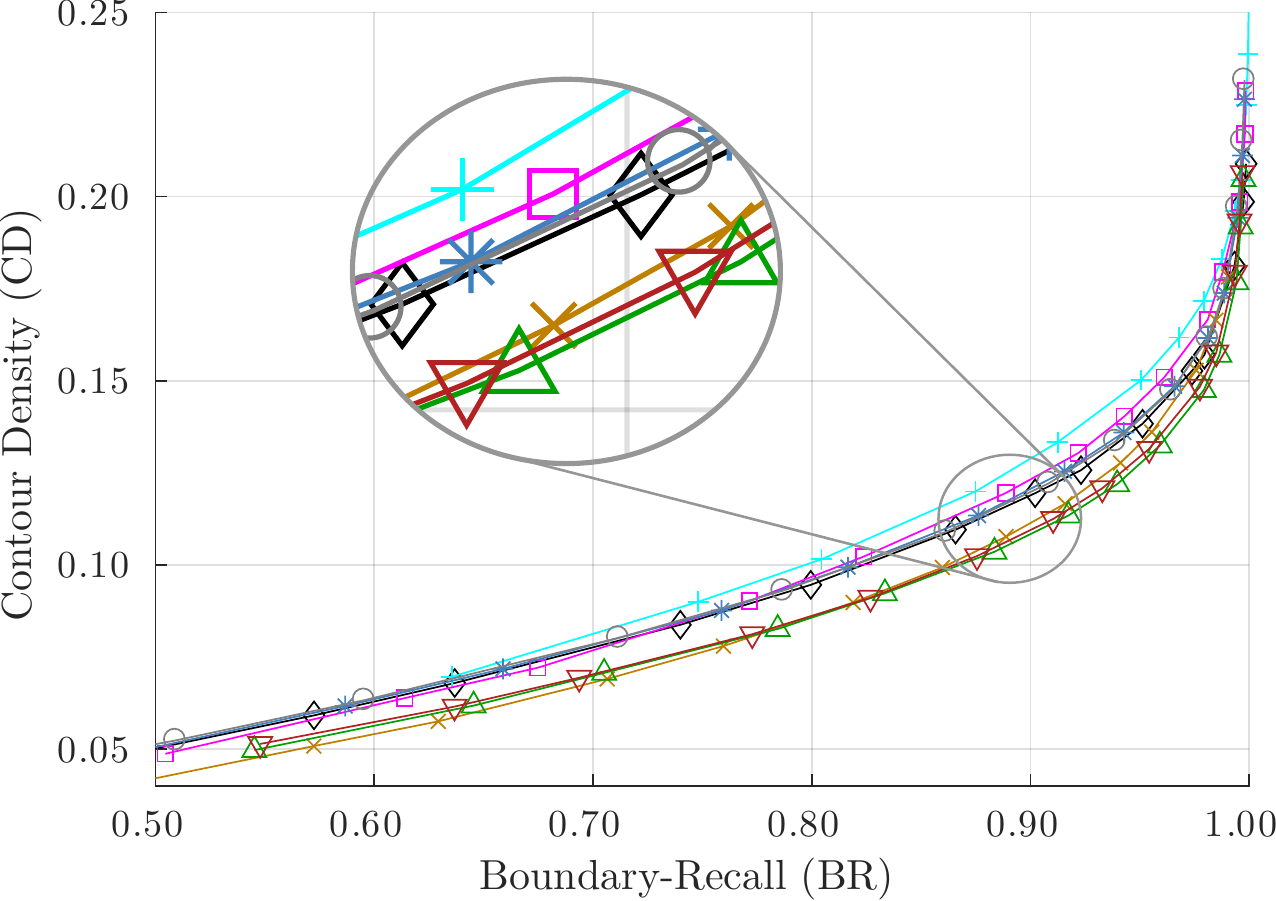}&
\includegraphics[width=\wwh,height=\hhh]{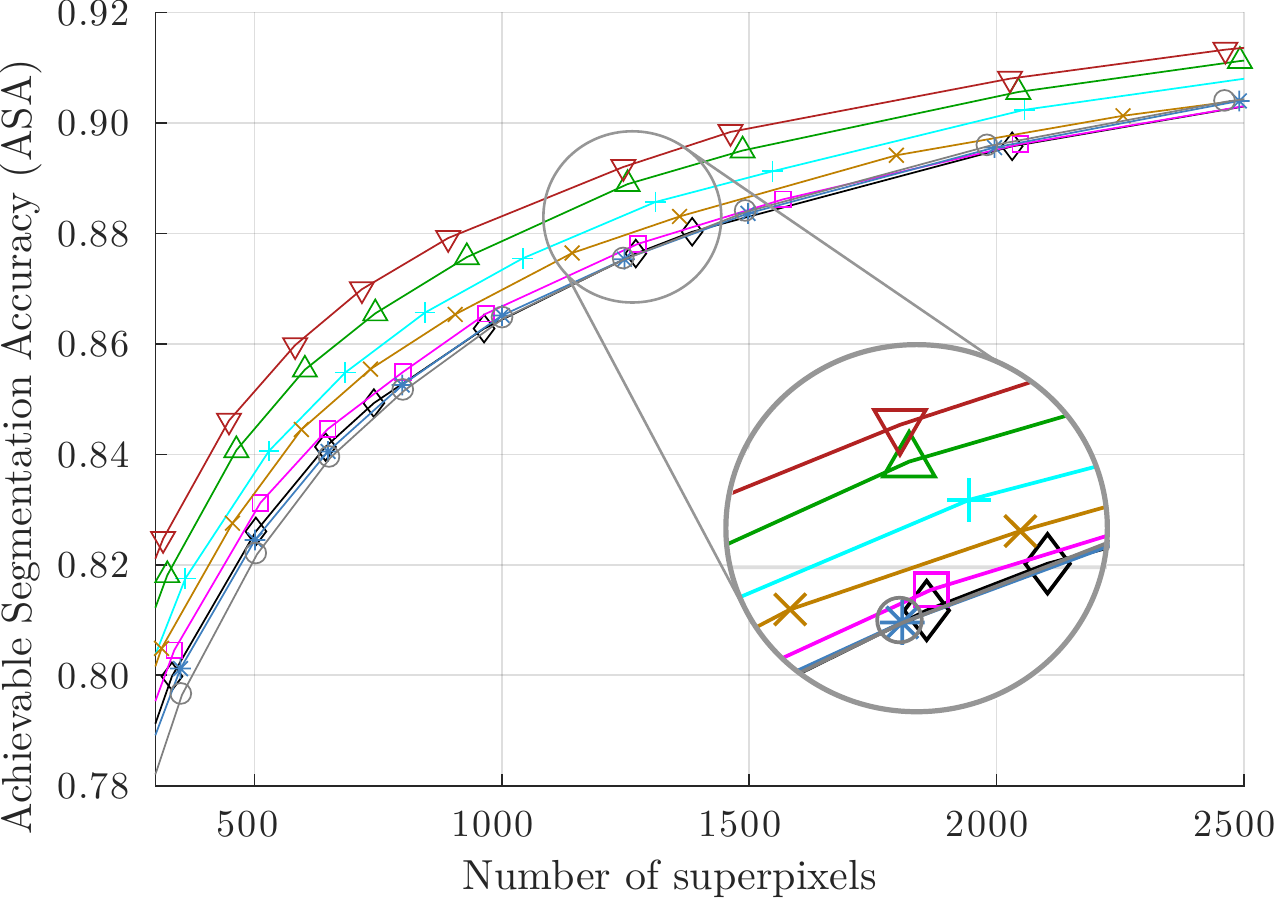}&
\includegraphics[width=\wwh,height=\hhh]{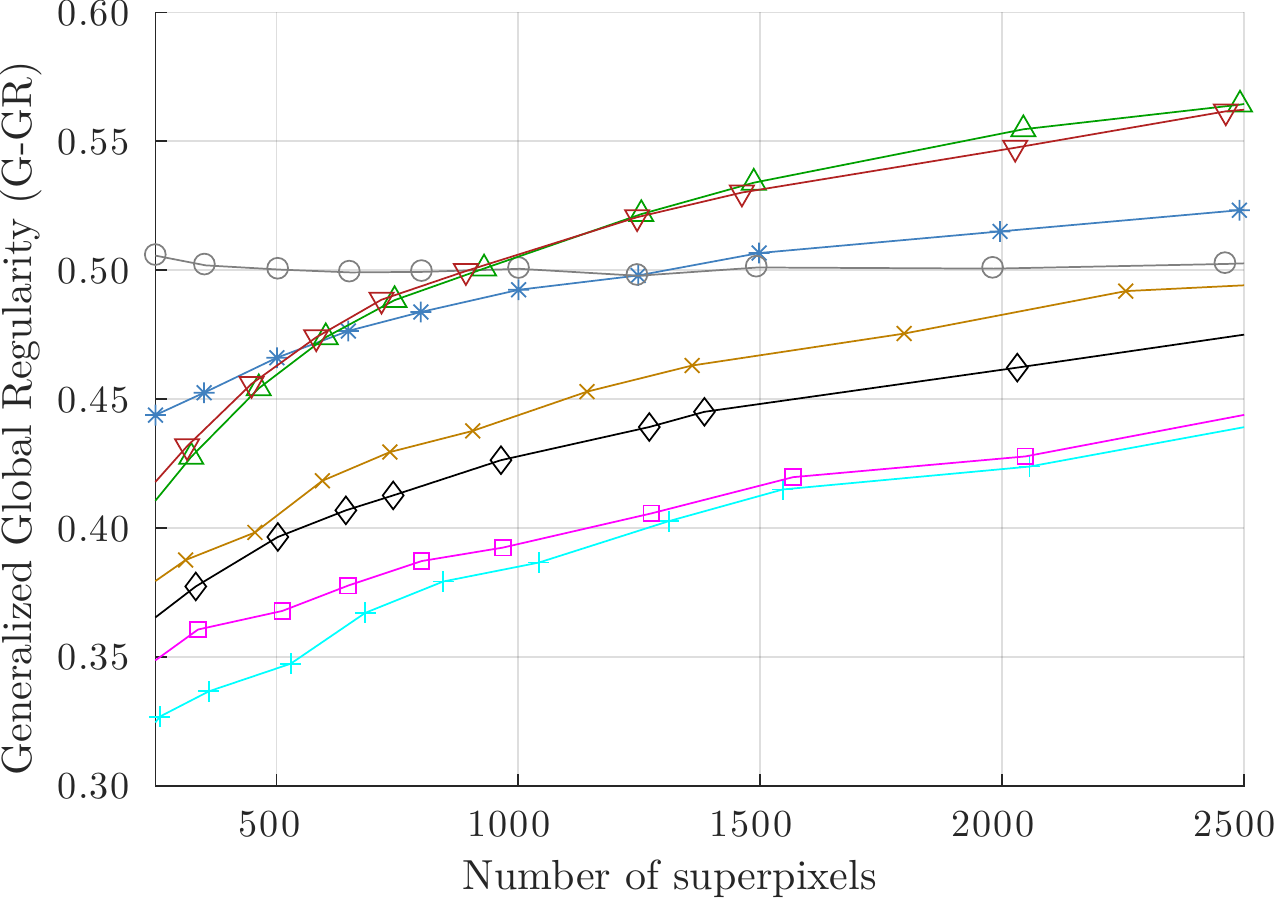}\\
\end{tabular}
} 
\caption{
Quantitative comparison on PR/BR, BR/CD, ASA and G-GR of the proposed SphSPS method to the state-of-the-art ones on the PSD \cite{zhao2018}.}%
\label{fig:sps_soa}
\end{figure*}

In Figure \ref{fig:sps_soa}, 
we report
the contour detection results measured by PR/BR 
curves 
with F-measure \eqref{fmeasure}, 
and BR/CD \eqref{br}, 
the segmentation of objects with  ASA \eqref{asa}, 
and regularity with the proposed G-GR metric \eqref{grs}, 
obtained for several numbers of superpixels.
%
%
SphSPS overall obtains the best segmentation results, 
with for instance the higher F-measure ($0.776$), 
and significantly outperforms the other spherical method SphSLIC, 
in both distance modes,
while producing very regular superpixels.
Note that even without the contour prior ($\gamma=0$), \emph{i.e.}, only using
color information on the shortest path, 
SphSPS still significantly outperforms the other state-of-the-art methods.
We also observe that by using a linear path approach, SCALP \cite{giraud2018_scalp} degrades the segmentation accuracy of LSC \cite{chen2017}.
This result highlights the need for considering our spherical shortest path instead of the linear one.

The regularity measured with the proposed G-GR \eqref{grs} appears to be very relevant
and able to differentiate planar and spherical methods.
It evaluates the convexity and contour smoothness of each superpixel along with their consistency, 
while COM \eqref{circu} is only based on a non robust and independent circularity assumption. 
Hence, with G-GR, 
the regularity in the spherical space is accurately measured such that
no planar methods have higher regularity than spherical ones
for a given number of superpixels, contrary to COM \cite{zhao2018}.

%
%
%
%

In Figure \ref{fig:sps_soa_img}, we show segmentation examples of SphSPS compared to the state-of-the-art methods, on 360$^\text{o}$ equirectangular images and projected on the unit sphere.
SphSPS produces regular superpixels in the spherical space and accurately captures the object contours compared to the other methods.

 Finally, in terms of processing time,
 the relevance of our features 
enables SphSPS to rapidly converge in a low number of iterations.
For instance, only using the 6 dimensional feature space \cite{chen2017},
SphSPS generates superpixels in $0.85$s per image of size $512{\times}1024$ pixels and already obtains higher accuracy  ($\text{F}=0.764$)
 than the state-of-the-art methods
  (see Figure \ref{fig:sps_param_curves}).
With the significant optimizations proposed in Section \ref{subsec:sps_path},
SphSPS can use the information on the shortest path to obtain significantly higher accuracy in only $2.30$s,
\emph{i.e.}, faster than existing spherical approaches \cite{zhao2018}.
Moreover, with basic multi-threading, 
we easily reduce the processing time of our implementation 
to $0.7$s to further facilitate the use of SphSPS\footnote{Available code at: \url{https://github.com/rgiraud/sphsps}}.

\newcommand{\ww}{0.23\textwidth} 
\newcommand{\ppp}{0.114\textwidth}  
\begin{figure*}[ht!]
\centering
{\scriptsize
\begin{tabular}{@{\hspace{1mm}}c@{\hspace{1mm}}c@{\hspace{1mm}}c@{\hspace{1mm}}c@{\hspace{3mm}}c@{\hspace{1mm}}c@{\hspace{1mm}}c@{\hspace{1mm}}c@{\hspace{0mm}}}
\rotatebox{90}{\hspace{0.55cm}LSC \cite{chen2017}}&
\includegraphics[width=\ww]{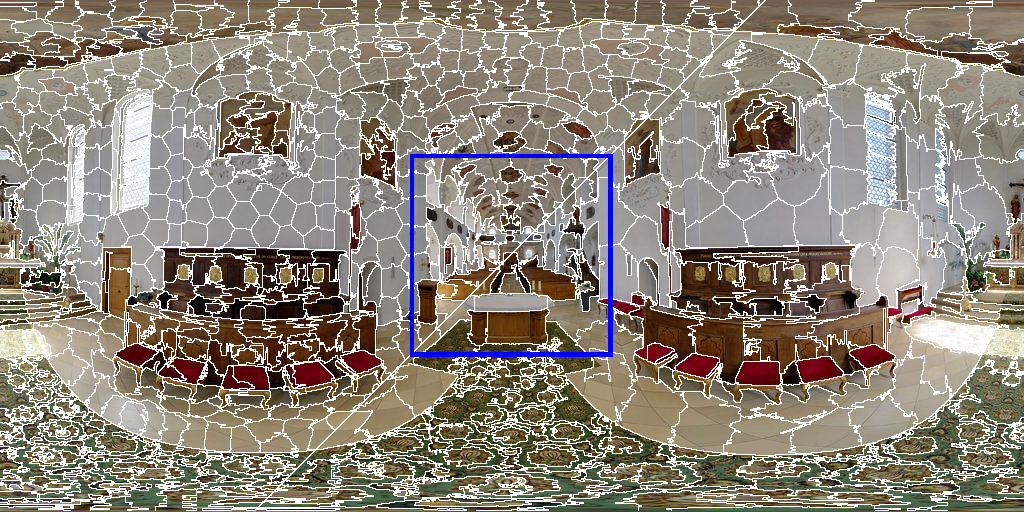}&
\includegraphics[width=\ppp,height=\ppp]{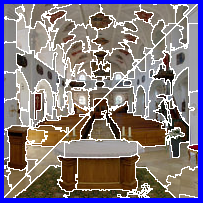}&
\includegraphics[width=\ppp,height=\ppp]{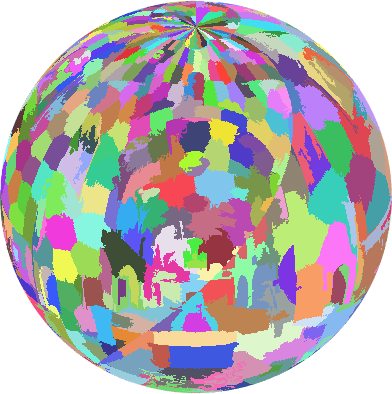}&
\rotatebox{90}{\hspace{0.0cm} SphSLIC-Euc \cite{zhao2018}}&
\includegraphics[width=\ww]{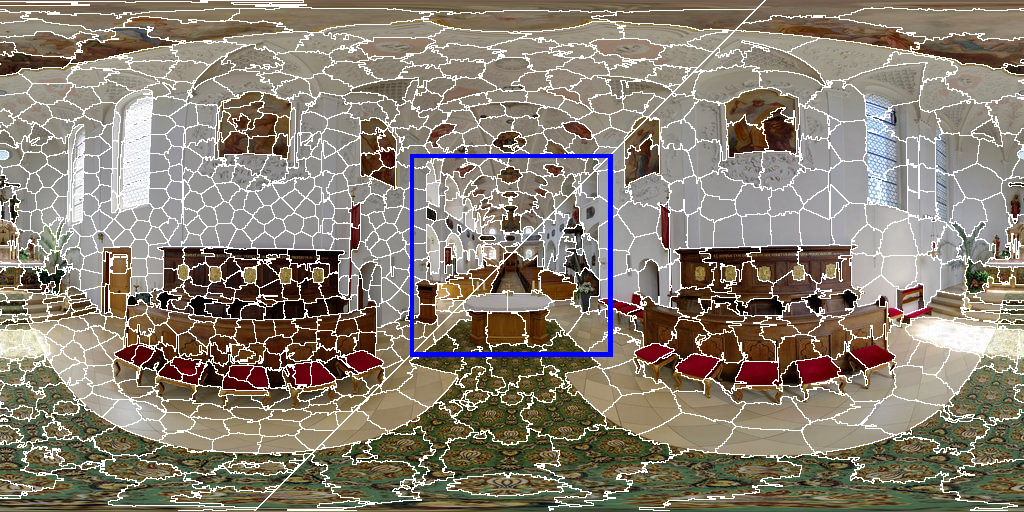}&
\includegraphics[width=\ppp,height=\ppp]{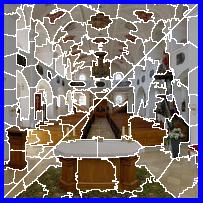}&
\includegraphics[width=\ppp,height=\ppp]{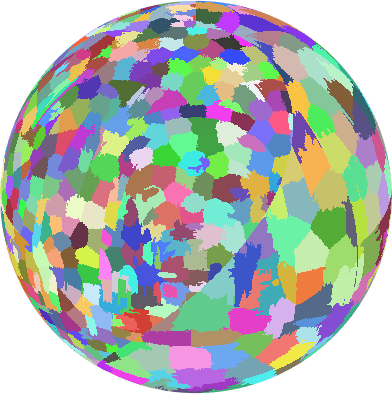}\\ 
\rotatebox{90}{\hspace{0.5cm}SNIC \cite{achanta2017superpixels}}&
\includegraphics[width=\ww]{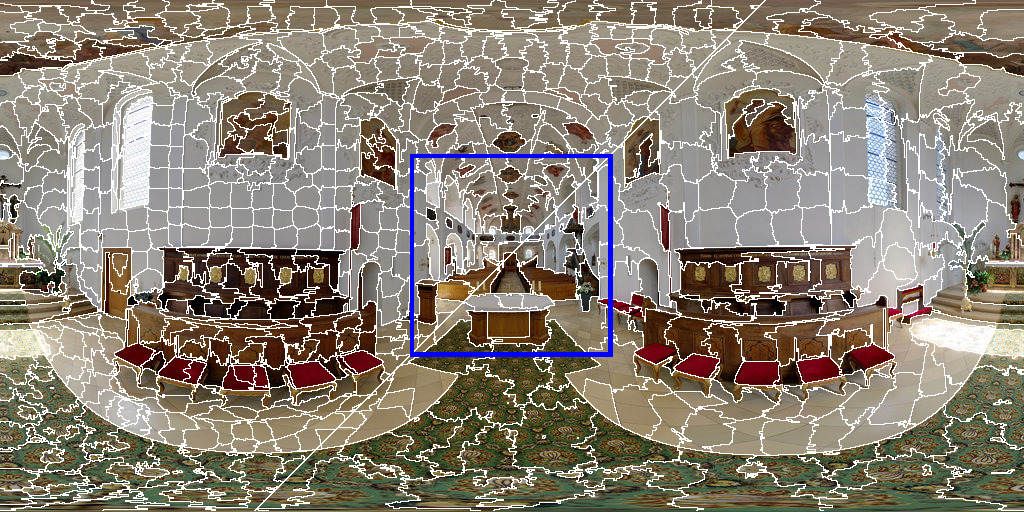}&
\includegraphics[width=\ppp,height=\ppp]{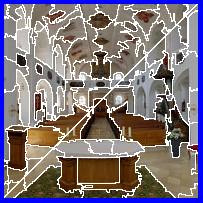}&
\includegraphics[width=\ppp,height=\ppp]{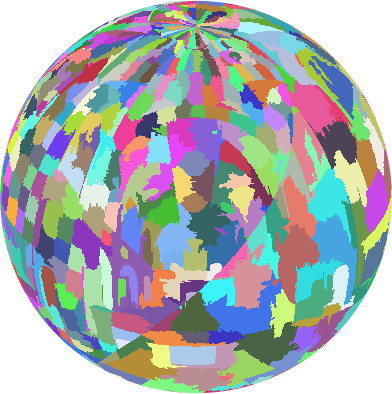}&
\rotatebox{90}{\hspace{0.00cm} SphSLIC-Cos \cite{zhao2018}} &
\includegraphics[width=\ww]{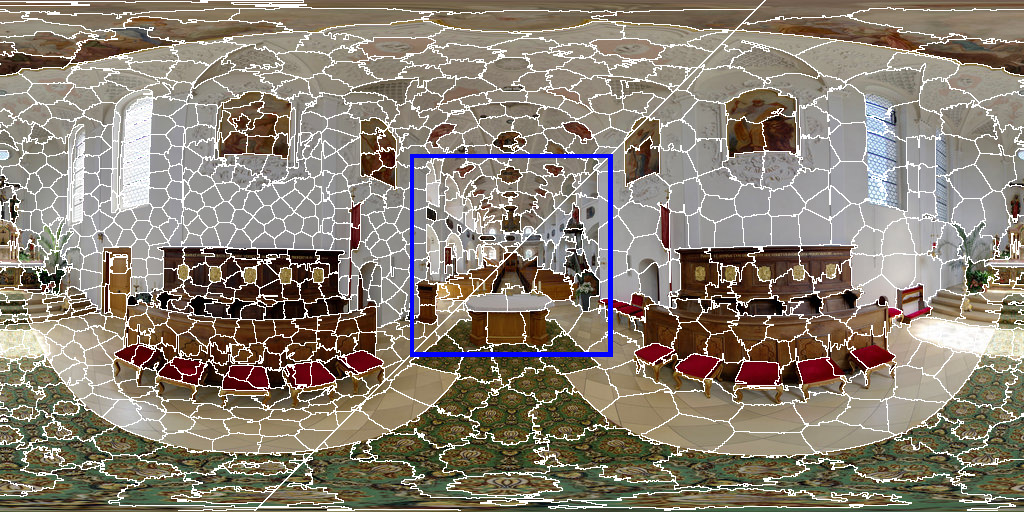}&
\includegraphics[width=\ppp,height=\ppp]{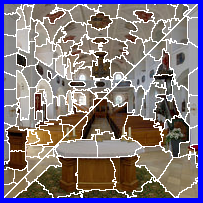}&
\includegraphics[width=\ppp,height=\ppp]{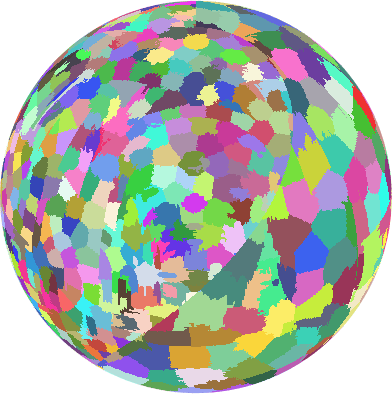}\\ 
\rotatebox{90}{\hspace{0.25cm} SCALP \cite{giraud2018_scalp}}&
\includegraphics[width=\ww]{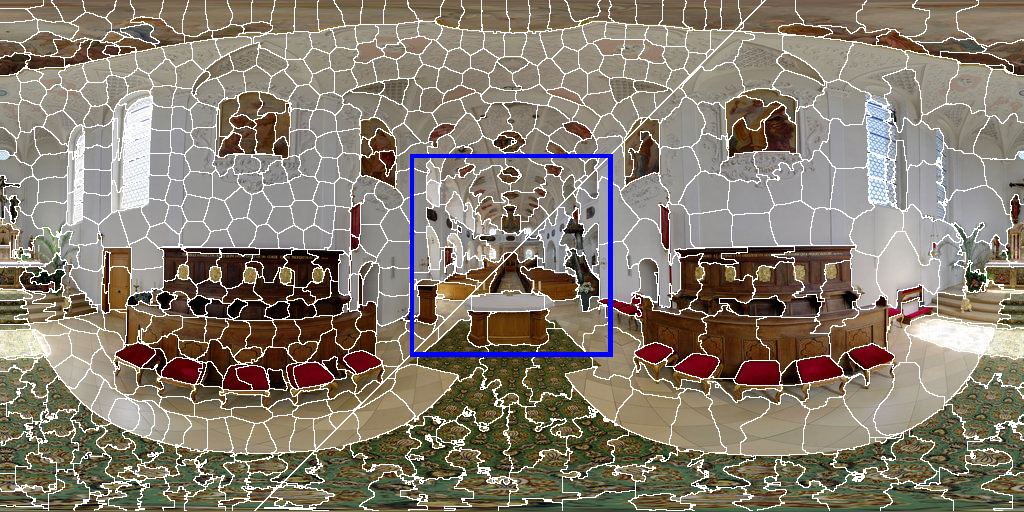}&
\includegraphics[width=\ppp,height=\ppp]{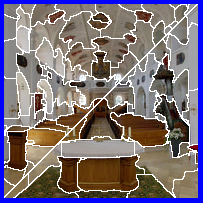}&
\includegraphics[width=\ppp,height=\ppp]{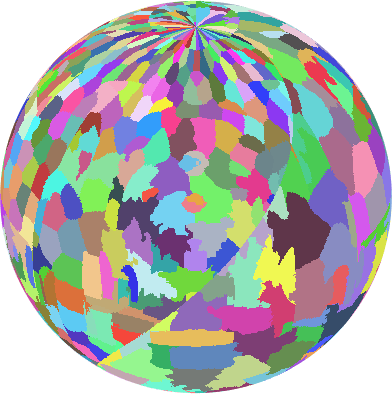}&
\rotatebox{90}{\hspace{0.75cm}\textbf{SphSPS}}&
\includegraphics[width=\ww]{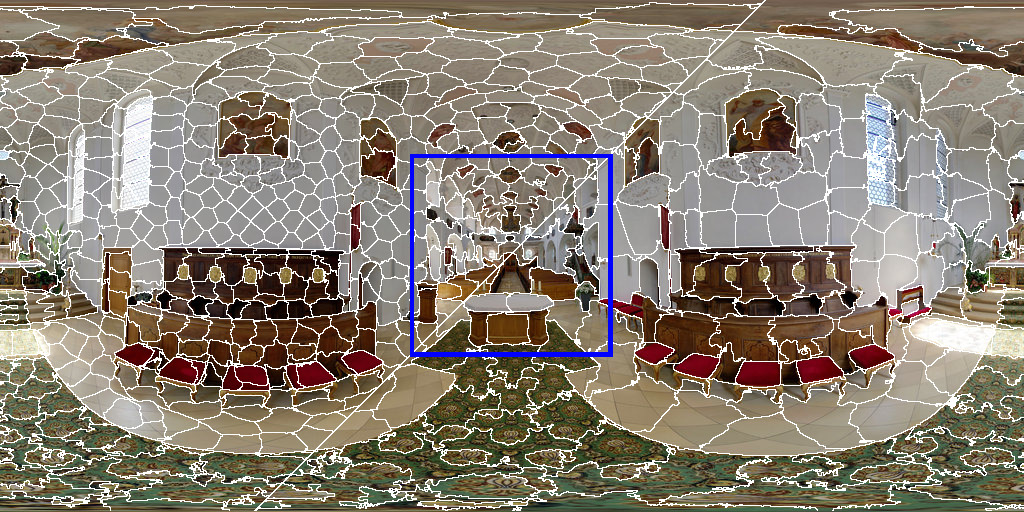}&
\includegraphics[width=\ppp,height=\ppp]{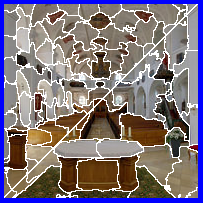}&
\includegraphics[width=\ppp,height=\ppp]{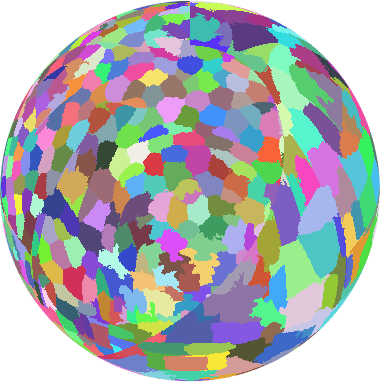}\\[0.25ex]
\rotatebox{90}{\hspace{0.55cm}LSC \cite{chen2017}}&
\includegraphics[width=\ww]{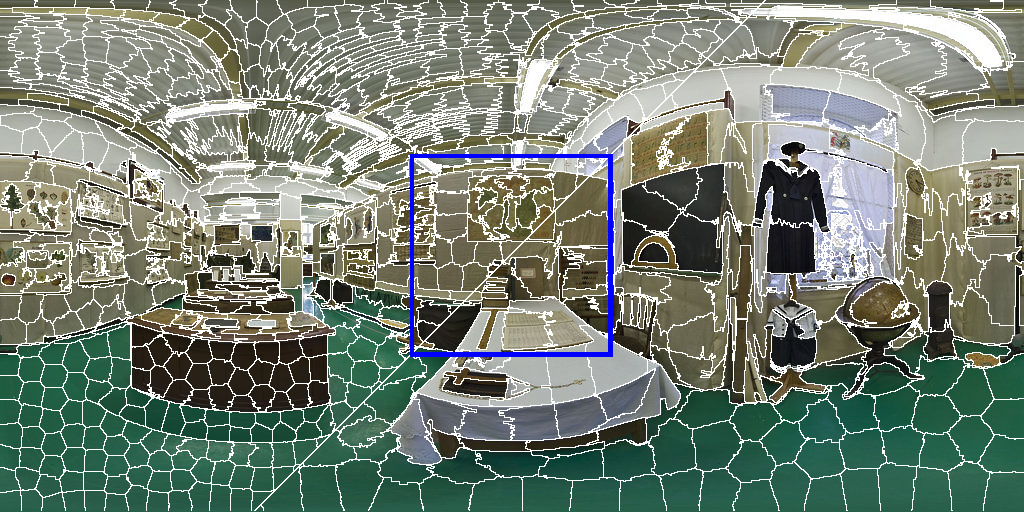}&
\includegraphics[width=\ppp,height=\ppp]{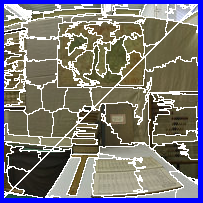}&
\includegraphics[width=\ppp,height=\ppp]{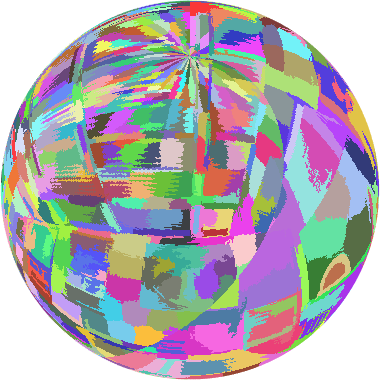}&
\rotatebox{90}{\hspace{0.0cm} SphSLIC-Euc \cite{zhao2018}}&
\includegraphics[width=\ww]{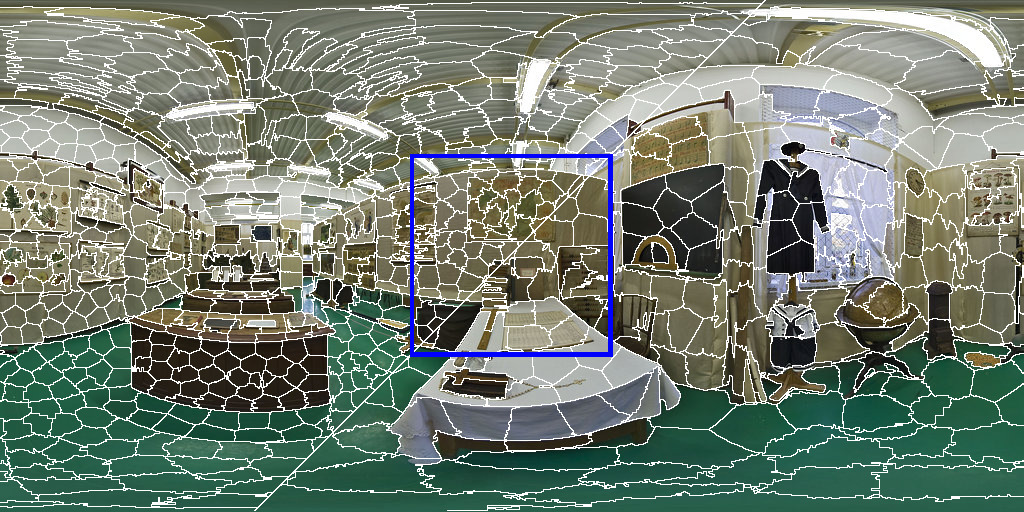}&
\includegraphics[width=\ppp,height=\ppp]{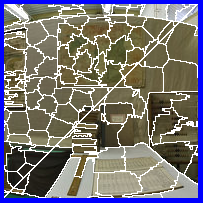}&
\includegraphics[width=\ppp,height=\ppp]{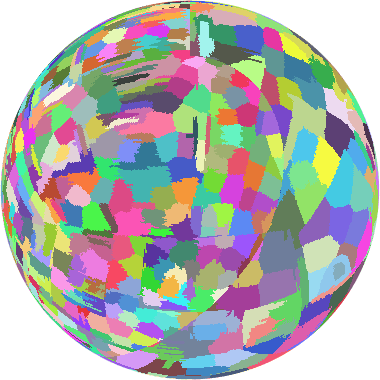}\\ 
\rotatebox{90}{\hspace{0.5cm}SNIC \cite{achanta2017superpixels}}&
\includegraphics[width=\ww]{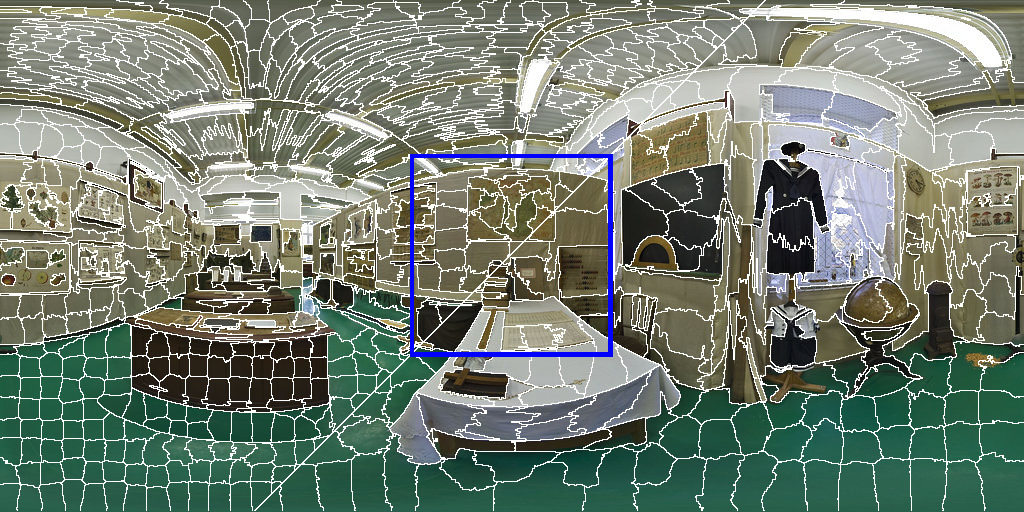}&
\includegraphics[width=\ppp,height=\ppp]{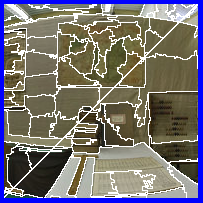}&
\includegraphics[width=\ppp,height=\ppp]{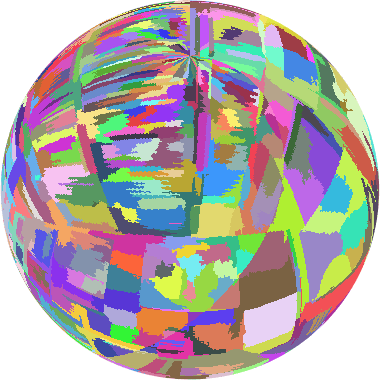}&
\rotatebox{90}{\hspace{0.00cm} SphSLIC-Cos \cite{zhao2018}} &
\includegraphics[width=\ww]{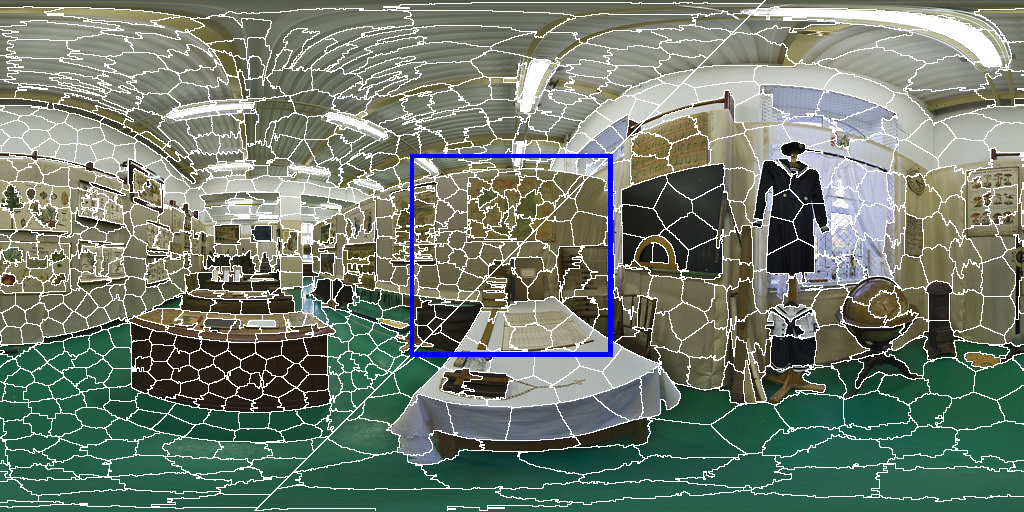}&
\includegraphics[width=\ppp,height=\ppp]{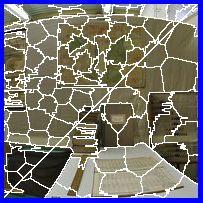}&
\includegraphics[width=\ppp,height=\ppp]{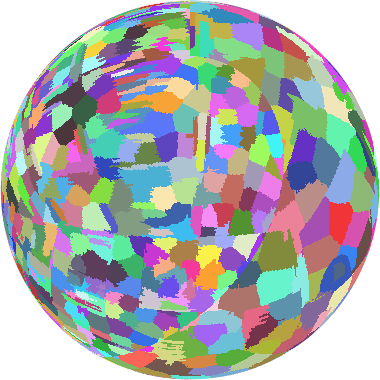}\\ 
\rotatebox{90}{\hspace{0.25cm} SCALP \cite{giraud2018_scalp}}&
\includegraphics[width=\ww]{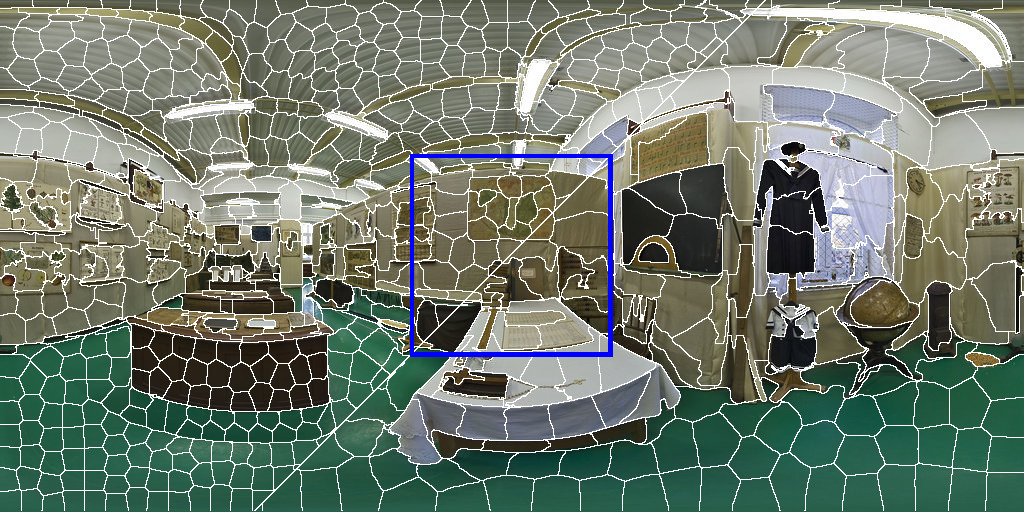}&
\includegraphics[width=\ppp,height=\ppp]{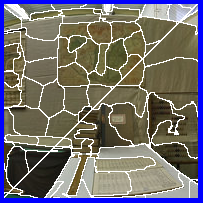}&
\includegraphics[width=\ppp,height=\ppp]{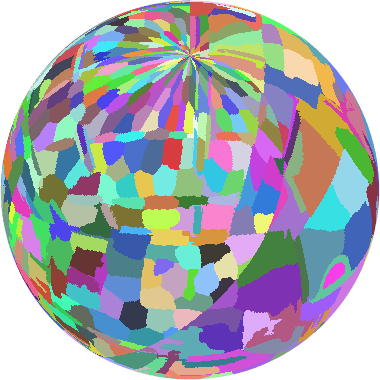}&
\rotatebox{90}{\hspace{0.75cm}\textbf{SphSPS}}&
\includegraphics[width=\ww]{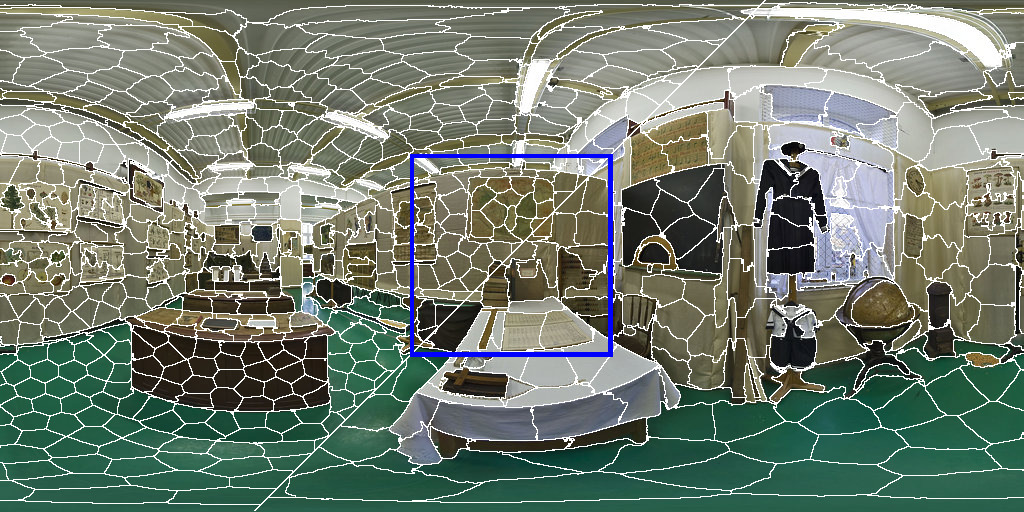}&
\includegraphics[width=\ppp,height=\ppp]{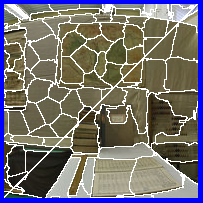}&
\includegraphics[width=\ppp,height=\ppp]{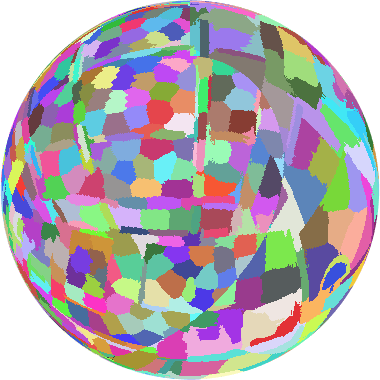}\\ 
\end{tabular}
} 
\caption{
Visual comparison between SphSPS and the 
best planar (left) and spherical (right) state-of-the-art 
methods on PSD images, for two superpixel numbers $K=1200$ (top-left) and $K=400$ (bottom right).
The compared methods may generate inaccurate superpixels, while
  SphSPS produces regular spherical superpixels with smooth boundaries
  that adhere well to the image contours.}%
\label{fig:sps_soa_img}
\end{figure*}

\section{Conclusion}

In this work, we generalize the shortest path approach
between a pixel and a superpixel barycenter \cite{giraud2017_jei} to the case of spherical images.
We show that the complexity resulting from the large number 
of pixels to process can be extremely 
reduced using the path redundancy on the 3D sphere.
Color features on this path
enable to generate both very accurate and regular superpixels.
Moreover, SphSPS can consider a contour prior information
to further improve its performances.

To ensure a relevant evaluation of regularity, 
we introduce a generalized metric 
measuring the spatial convexity and consistency in the 3D spherical space.
While providing regular results in the acquisition space, 
SphSPS significantly outperforms both planar and spherical state-of-the-art methods.

Accuracy and regularity in the acquisition 
space are crucial for relevant 
display and 
for computer vision pre-processing.
Future works will extend our method to spherical videos and other 
acquisition spaces, \emph{e.g.}, circular or polarimetric.

\bibliographystyle{IEEEtran}
\bibliography{biblio}

\end{document}